\documentclass[11pt]{article}

\usepackage[preprint]{acl}

\usepackage{times}
\usepackage{latexsym}
\usepackage[T1]{fontenc}
\usepackage[utf8]{inputenc}
\usepackage{microtype}
\usepackage{inconsolata}
\usepackage{graphicx}

\usepackage{mathtools}
\usepackage{booktabs}
\usepackage{colortbl}
\usepackage{multirow}
\usepackage{amsmath}
\usepackage{amssymb}
\usepackage{float}

\definecolor{Red}{rgb}{0.768, 0.054, 0.054}
\definecolor{Blue}{rgb}{0.152, 0.294, 0.925}
\definecolor{Green}{rgb}{0,0.4,0.7}
\definecolor{rulemethodrow}{RGB}{242,244,247}
\definecolor{llmmethodrow}{RGB}{232,246,239}
\hypersetup{
    colorlinks=true,
    citecolor=teal,
    linkcolor=Red,
    urlcolor=Green,
}

\usepackage{tcolorbox}
\tcbuselibrary{skins} 
\tcbuselibrary{listings}
\tcbuselibrary{breakable}
\newtcblisting{promptbox}[1][]{
    enhanced,
    attach boxed title to top left={xshift=1em, yshift=-\tcboxedtitleheight/2},
    coltitle=white,
    width=\textwidth,
    listing only,
    boxed title style={sharp corners, frame hidden},
    fonttitle=\ttfamily\centering\bfseries,
    title={#1},            
    listing options={
        language={},
        basicstyle=\tiny\ttfamily,
        breaklines=true,
        breakautoindent=false,
        breakindent=0pt,
        columns=fullflexible,
        showstringspaces=false,
    },
    boxsep=2pt,
    top=2pt,
    bottom=2pt,
}

\definecolor{targettaskfill}{RGB}{247,249,252}
\definecolor{targettaskframe}{RGB}{104,116,130}
\definecolor{targettrajfill}{RGB}{255,248,248}
\definecolor{targettrajframe}{RGB}{190,45,45}
\definecolor{targetfeedbackfill}{RGB}{246,250,255}
\definecolor{targetfeedbackframe}{RGB}{42,92,170}
\definecolor{targettoolcall}{RGB}{36,91,168}
\definecolor{targettoolresult}{RGB}{0,122,96}
\tcbset{
    targettaskstyle/.style={
        enhanced,
        breakable,
        colback=targettaskfill,
        colframe=targettaskframe,
        colbacktitle=targettaskframe,
        coltitle=white,
        fonttitle=\bfseries,
        boxrule=0.6pt,
        arc=1mm,
        left=5pt,
        right=5pt,
        top=5pt,
        bottom=5pt,
    },
    targettrajstyle/.style={
        enhanced,
        breakable,
        colback=targettrajfill,
        colframe=targettrajframe,
        colbacktitle=targettrajframe,
        coltitle=white,
        fonttitle=\bfseries,
        boxrule=0.6pt,
        arc=1mm,
        left=5pt,
        right=5pt,
        top=5pt,
        bottom=5pt,
    },
    targetfeedbackstyle/.style={
        enhanced,
        breakable,
        colback=targetfeedbackfill,
        colframe=targetfeedbackframe,
        colbacktitle=targetfeedbackframe,
        coltitle=white,
        fonttitle=\bfseries,
        boxrule=0.6pt,
        arc=1mm,
        left=5pt,
        right=5pt,
        top=5pt,
        bottom=5pt,
    },
}

\newcommand{\appref}[1]{\hyperref[#1]{Appendix~\ref*{#1}}}

\title{EvolveTrade: Experience-Driven Policy Refinement\\ for Self-Evolving LLM Trading Agents}

\author{
Sehee Kim$^{*\,1}$ \quad Yumin Choi$^{*\,1}$ \quad Minki Kang$^{1}$ \quad Sung Ju Hwang$^{1, 2}$ \\[4pt]
$^{1}$KAIST \quad $^{2}$DeepAuto.ai \\[4pt]
\texttt{\{sehee.kim, yuminchoi, sungju.hwang\}@kaist.ac.kr}
}

\newcommand\blfootnote[1]{%
  \begingroup
  \renewcommand\thefootnote{}\footnote{#1}%
  \addtocounter{footnote}{-1}%
  \endgroup
}

\begin{document}
\maketitle


\blfootnote{\textsuperscript{*}Equal contribution.}

\begin{abstract}
Large language model (LLM) trading agents can combine market data, news, and executable analysis, but their behavior is often controlled by static hand-written tool-use policies that are fixed before deployment.
This limits their ability to adapt how they gather evidence, invoke tools, verify signals, and manage risk under changing market regimes.
We introduce \textbf{EvolveTrade}, a self-evolving framework that treats the system prompt of a tool-using trading agent as a text-parameterized policy.
After each update interval, a Policy Agent revises this policy using accumulated decision traces and realized portfolio feedback, while keeping the backbone LLM fixed.
The updated policy is then used for the next batch of trading decisions, enabling the agent to refine its information-acquisition and portfolio-construction procedure over time.
Experiments across multiple market regimes and two LLM backbones show that EvolveTrade often improves Sharpe Ratio and Cumulative Return over fixed-policy LLM baselines, achieving the improved SR and CR in most evaluated settings.
Behavioral analyses further show that self-evolved policies increase code-mediated analysis and activate regime-relevant computations; case-level policy-to-return attributions trace how policy-induced allocation changes contribute to realized return differences.
These results suggest that adapting the reusable procedure governing tool use is a key direction for building more robust LLM trading agents.
\end{abstract}

\section{Introduction}
Large Language Models (LLMs) have demonstrated strong capabilities in reasoning, instruction following, and integrating heterogeneous information sources~\citep{gpt5, qwen35}.
Yet these capabilities do not reliably translate into robust decision-making in financial markets~\citep{livetradebench}, where an agent must repeatedly acquire relevant information, verify noisy and conflicting signals, manage portfolio risk, and act under delayed feedback in a non-stationary environment~\citep{non-stationarity}.

Recent work has explored LLMs as trading agents by leveraging their ability to combine numerical market data with unstructured qualitative information, such as financial news, market sentiment, and analyst reports, under natural language instructions~\citep{tradingagents}.
Other studies further equip agents with memory, planning, and external tools, allowing them to interact with market data APIs, technical indicator libraries, news search engines, and trading interfaces~\citep{ReAct,FinMem,AITrader}.
These systems show that LLM agents can support more flexible financial decision-making than purely numerical models.
However, they also suggest that richer context or tool access alone is insufficient: \emph{the agent must know when, why, and how to use these resources.}

\begin{figure*}[t]
    \centering
    \includegraphics[width=\textwidth]{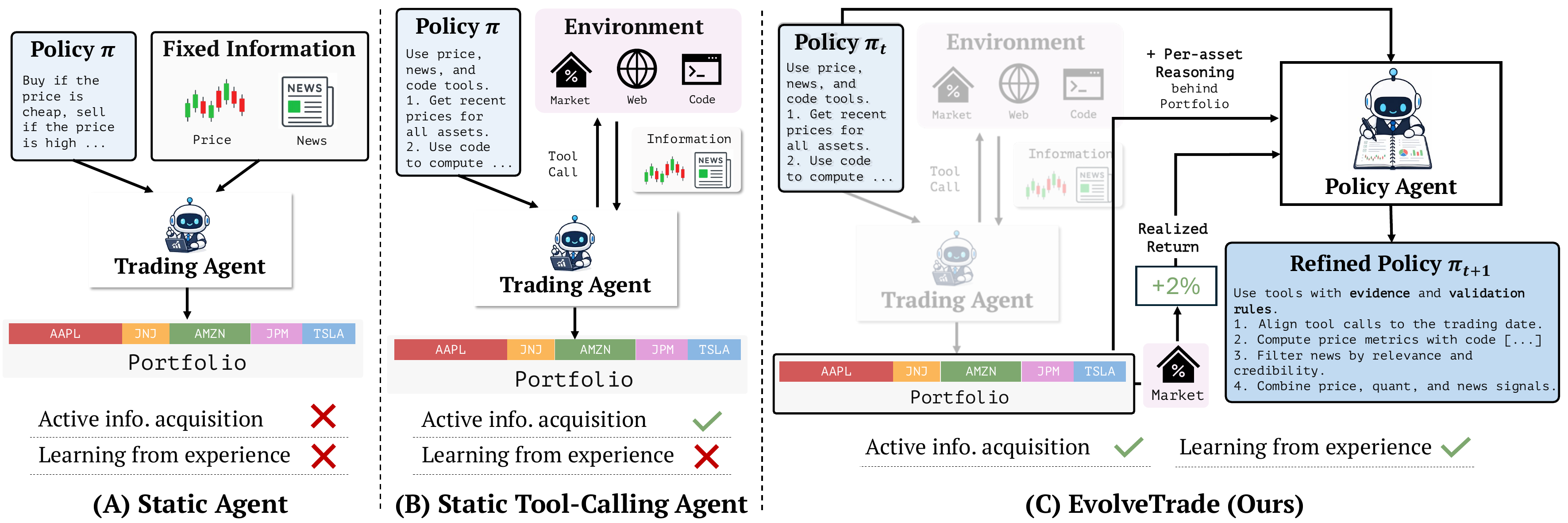}
    \vspace{-0.25in}
    \caption{
    \textbf{Concept.}
    \textbf{(A)} The \emph{Static Agent} acts on a fixed observation channel of price and news, with no tool use and no learning from feedback.
    \textbf{(B)} The \emph{Static Tool-Calling Agent} actively retrieves information using price, news, and code tools, but its policy is fixed at deployment.
    \textbf{(C)} \textbf{EvolveTrade (ours)} adds a Policy Agent that, after each completed trading interval, reads the realized trade together with the agent's per-asset reasoning and refines the policy, so the policy on any later trading day has been shaped by the agent's own realized experience.}
    \label{fig:concept}
    \vspace{-0.15in}
\end{figure*}

Existing approaches typically fix this information acquisition procedure at development time, in two broad forms as summarized in \autoref{fig:concept}.
Live or realistic trading benchmarks~\citep{livetradebench, QuantAgent, tradingagents} provide agents with a predefined observation window of prices, portfolio states, and market news, and evaluate how well the agent decides from the given information (\autoref{fig:concept}(A)).
Autonomous trading frameworks~\citep{AITrader} allow agents to call external tools, but their tool-use behavior is still governed by fixed prompts, predefined workflows, or static agent roles (\autoref{fig:concept}(B)).
In both cases, the decision-making procedure remains a pre-defined prompt that is never revised by realized trading.

We argue that this fixed policy can be a central bottleneck for LLM trading agents.
The right object to shape by experience is the agent's reusable procedure for acquiring, validating, and acting on information, rather than any specific trading decision.
We therefore need a mechanism that lets the agent refine that procedure between trading days, so that past experience informs today's decision.

We introduce \textbf{EvolveTrade}, a self-evolving framework in which the LLM refines its own tool-use policy from its own trading experience (\autoref{fig:concept} (C)).
After each completed trading interval, a separate Policy Agent reads the realized trade together with the agent's per-asset reasoning and refines the policy text from $\pi_t$ to $\pi_{t+1}$, so the policy active on any later trading day has been shaped by the agent's own realized experience.
A single refinement step can change which tools the agent prioritizes, which signals it cross-validates against each other, or how it adjusts exposure under specific market conditions, while the underlying LLM parameters and tool interfaces are left untouched throughout.
The only object EvolveTrade learns is the policy text itself, so the framework can be applied to any frozen LLM trading agent without retraining the model or modifying its tools.

We empirically validate two findings.
First, online policy self-evolution often improves LLM trading agents over fixed-policy baselines: across two backbones and six post-cutoff market regimes, EvolveTrade achieves the best Sharpe Ratio and Cumulative Return among LLM-based methods in three model-regime pairs, while fixed-policy agents remain stronger in the other two.
Second, the improvement is accompanied by measurable changes in tool-use behavior.
EvolveTrade increases code-mediated analysis, activates regime-relevant computations that the Static Tool-Calling Agent never invokes, including VaR and signal normalization in the April drawdown and trend-following indicators such as EMA, RSI, and SMA in the September uptrend, and in a January 2025 NVDA drawdown case its refined sizing policy accounts for a +1.33 percentage-point relative daily return difference.
These results suggest that LLM trading agents should be allowed to revise the procedure by which they use their tools from their own trading experience.

\section{Related Work}

\paragraph{LLM Agents for Financial Trading.}
Recent work has explored LLMs as financial trading agents by augmenting them with memory, tools, multimodal signals, and multi-agent deliberation. FinMem~\citep{FinMem} introduces layered memory for retaining market observations across sessions, while TradingAgents, StockAgent, QuantAgent, and TradExpert~\citep{tradingagents,stockagent,QuantAgent,tradexpert} organize decision making through specialized agent roles, simulated trading environments, or fixed analysis pipelines. FinAgent and related financial agent platforms~\citep{multimodalfoundationagent,finsphere,finrobot} incorporate multimodal inputs, real-time data, domain tools, or reflective decision making, and recent benchmarks such as AI-Trader, LiveTradeBench, and FinAgentBench~\citep{AITrader,livetradebench,finagentbench} evaluate LLM agents under more realistic financial information and market conditions. Recent studies also examine reliability issues in LLM trading, including noisy-source trust, spurious ticker memorization, and temporal leakage~\citep{trusttrade,blindtrade,lookaheadbench}. These systems show that LLMs can integrate heterogeneous financial evidence, but their observation channels, prompts, role structures, or tool workflows are typically specified before deployment. Thus, the operational policy governing when an agent should acquire information, verify evidence, and commit to an allocation remains largely static.

\paragraph{Self-evolving LLM Agents.}
A related line of work studies how LLM agents can improve from their own execution traces, feedback, or task outcomes. This direction builds on tool-using agents that interleave reasoning with external actions or learn when to invoke APIs~\citep{ReAct,toolformer}, but shifts the focus from fixed tool access to post-deployment adaptation. General prompt, pipeline, and workflow optimization methods revise instructions or LM programs from validation signals, gradient-like feedback, or evolutionary search~\citep{protegi,opro,dspy,textgrad,gepa,MetaSPO,aflow}. Agent-level self-evolution methods further convert successes and failures into reflective memories, reusable workflows, or reasoning strategies~\citep{reflexion,voyager,awm,reasoningbank,skillmas}. Within trading, ATLAS and SHARP adapt prompts or structured policies from market feedback~\citep{atlas,sharp}, and AlphaQuanter and FLAG-Trader~\citep{alphaquanter,flagtrader} show that learned tool orchestration or policy optimization can improve over fixed multi-agent baselines. However, these methods either optimize reasoning instructions, maintain experience memories, or learn tool-use policies offline, rather than continuously refining the operational procedure by which a deployed agent gathers, validates, and acts on information. EvolveTrade instead treats the system prompt as a text-parameterized tool-use policy and updates it online from the agent's own tool-use traces, trading decisions, and portfolio feedback.

\section{Problem Setup: Tool-Using Trading Agent}
\label{sec:setting}

We consider a sequential trading setting over trading days $t=1,\ldots,T$.
At the beginning of day $t$, the Trading Agent is initialized with
\begin{equation}
    S_t = (\pi_t, t, P_{t-1}, \mathcal{T}),
\end{equation}
where $\pi_t$ is the tool-use policy in effect on day $t$, $P_{t-1}$ is the previous portfolio state, and $\mathcal{T}$ is the available toolset.
The policy is implemented as the natural-language system prompt that guides the agent's tool use, evidence verification, risk control, and output requirements.
We write $\pi_t$ to allow the policy text to vary across days.
In a baseline hand-crafted setting the policy is held constant ($\pi_t \equiv \pi_0$), and \autoref{sec:method} describes how EvolveTrade updates $\pi_t$ from realized trading experience.
The actual action distribution is induced jointly by the frozen LLM $p_\theta$, the current context, the available tools, and the policy text $\pi_t$.
The LLM parameters and tool interfaces are fixed throughout.

At decision time $t$, the agent can access only information available before the allocation is submitted.
Market information is not assumed to be pre-injected into the initial prompt, unlike static-observation settings such as LiveTradeBench~\citep{livetradebench}.
Instead, the agent actively retrieves and processes information through tools during its decision process.
In our implementation, the toolset is
$\mathcal{T} = \{t_\mathrm{price}, t_\mathrm{news}, t_\mathrm{code}\}$,
consisting of a price retrieval tool and a news search tool from recent autonomous trading frameworks~\citep{AITrader}, augmented by a Python code interpreter.
All retrieval tools enforce the same temporal cutoff to prevent look-ahead leakage.

The agent outputs a feasible portfolio allocation $w_t \in \mathcal{W}$,
where $\mathcal{W}$ denotes the portfolio constraint set.
In our experiments, $\mathcal{W}$ is a long-only allocation simplex over the tradable assets and cash unless otherwise specified.
The allocation $w_t$ is executed after the decision time and evaluated over the next holding interval~\citep{livetradebench}.

Conditioned on $\pi_t$, the frozen LLM $p_\theta$ runs an inner loop of tool invocations interleaved with reasoning steps, terminating when it emits the final allocation $w_t$ and a per-asset rationale $d_t^{\mathrm{dec}}$ that records, for each asset, the evidence and arguments the agent used to set its weight.
We denote the agent's recorded output for day $t$ as
\begin{equation}
    h_t = (w_t, d_t^{\mathrm{dec}}),
\end{equation}
which is the decision trace passed downstream to the policy-refinement step.

\begin{figure*}[t]
    \centering
    \includegraphics[width=\textwidth]{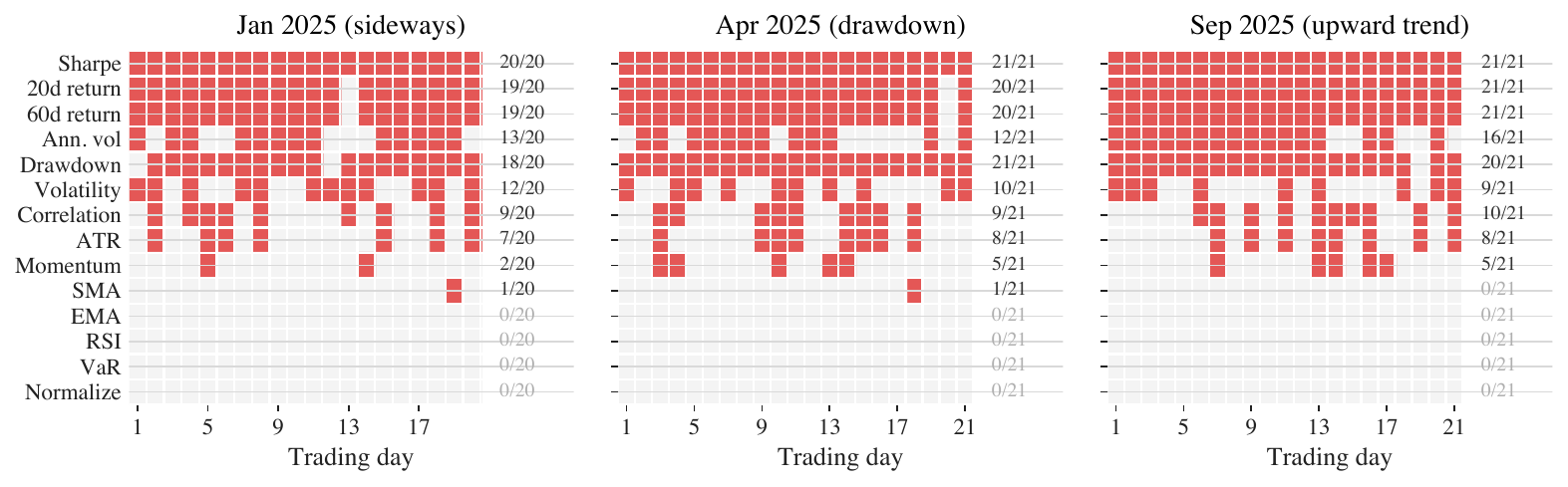}
    \vspace{-0.3in}
    \caption{
    \textbf{Static policy locks the LLM into a fixed analytic vocabulary.}
    Each panel is a per-day binary map of analytic metrics that appear in the agent's code-interpreter calls during the corresponding one-month period.
    Rows are ordered by overall usage frequency, and numbers on the right report how many of the period's trading days used each metric.
    The same five metrics light up every day in all three regimes, while standard regime-relevant metrics in the bottom rows remain almost unused.
    \autoref{fig:coverage-unlocked} shows how EvolveTrade activates these previously unused metrics.}
    \label{fig:motivation-coverage}
    \vspace{-0.2in}
\end{figure*}

\section{Motivation: Static Policy Locks the LLM into a Fixed Analytic Coverage}
\label{sec:motivation}

\paragraph{Setup.}
We instantiate the trading agent of \autoref{sec:setting} with a fixed policy ($\pi_t \equiv \pi_0$) and run it under \texttt{GPT-5-mini} across three one-month periods that span distinct market regimes: a sideways month (Jan.\ 2025), a sharp drawdown with V-shaped recovery (Apr.\ 2025), and a steady upward trend (Sep.\ 2025).
The full experimental setup is described in \autoref{sec:experiments}.

\paragraph{The analytic vocabulary of agent is regime-invariant.}
\autoref{fig:motivation-coverage} illustrates which analytic metrics the agent actually invokes inside its Python code interpreter calls, for each trading day in the three regimes.
The usage trend is essentially the same in every regime.
The same five metrics (Sharpe, 20d return, 60d return, annualized volatility, drawdown) appear on $90$--$100\%$ of days, as LLM follows the static tool-use policy mentioned in the prompt (\autoref{fig:prompt-evolvetrade-initial}). A handful of additional metrics (correlation, ATR, momentum) appear sporadically and standard regime-relevant computations such as RSI, EMA, VaR, and signal normalization never appear at all.
The same pattern holds for the other tools: news queries reduce to a per-asset boilerplate (e.g.\ \texttt{TICKER}, \texttt{TICKER news}), and the price-retrieval lookback is fixed at roughly six months on every call across all three regimes.

\paragraph{Why this motivates a self-evolving framework.}
The initial system prompt $\pi_0$ is the only signal that shapes which subset of the LLM's analytical capability gets invoked, and because that prompt is fixed before any market data has been observed, the subset is decided at development time and never revisited.
The agent applies the same analytical scope to a $-9.8\%$ drawdown (Apr 2025) as to a steady upward trend (Sep 2025), even though tail-risk metrics and trend-following indicators are clearly more informative in one of those regimes than the other.
More importantly, the agent has no way to improve from realized trading experience: by construction, nothing in the prompt is responsive to what its earlier trades revealed about which tools to invoke or which metrics to compute.
This motivates a framework in which the LLM revises its own tool-use procedure between trading days from its own trajectory and post-trade feedback, which we introduce as \textbf{EvolveTrade} in \autoref{sec:method}.

\section{EvolveTrade}
\label{sec:method}

\textbf{EvolveTrade} is a self-evolving framework for the tool-using LLM trading agent defined in \autoref{sec:setting}, in which the LLM rewrites its own policy from its own trading experience.
We refer to this online process as \emph{policy self-evolution}, and to each LLM-driven text update as a \emph{refinement step}.
We define the per-day refinement record in \autoref{sec:method:feedback} and the refinement step itself in \autoref{sec:method:refinement}.

\subsection{Trading Experience as Policy Feedback}
\label{sec:method:feedback}

After the agent submits the asset allocation $w_t$, the environment updates the portfolio state from $P_{t-1}$ to $P_t$ and returns post-trade feedback $g_t$.
We define $g_t$ as a feedback record summarizing the realized effect of the decision on day $t$, comprising the daily portfolio return, the per-asset returns of the tradable assets, and the resulting allocations and portfolio state.

However, $g_t$ alone is an ambiguous signal.
A loss may reflect flawed evidence gathering or an unavoidable market shock.
A gain may reflect sound analysis or favorable noise.
This ambiguity is especially important in financial markets, where delayed and noisy feedback makes direct policy optimization difficult~\citep{metatrader}.
EvolveTrade therefore pairs the decision trace $h_t$ with the post-trade feedback $g_t$ into a single refinement record: $\mathcal{R}_t = (h_t, g_t)$.
The decision trace exposes the allocation the agent emitted and the per-asset rationale that justified it, while the feedback supplies ex-post performance context.
EvolveTrade uses $\mathcal{R}_t$ to identify plausible procedural weaknesses and reusable safeguards for revising the policy.

\begin{table*}[t]
\centering
\small

\setlength{\tabcolsep}{3.0pt}
\resizebox{\textwidth}{!}{%
\begin{tabular}{lccccccccccccccc}
\toprule
& \multicolumn{5}{c}{\textbf{Jan. 2025}}
& \multicolumn{5}{c}{\textbf{Apr. 2025}}
& \multicolumn{5}{c}{\textbf{Sep. 2025}} \\
\cmidrule(lr){2-6}
\cmidrule(lr){7-11}
\cmidrule(lr){12-16}
\textbf{Methods}
& SR$\uparrow$ & CR\%$\uparrow$ & MDD\%$\downarrow$ & WR\%$\uparrow$ & Vol\%$\downarrow$
& SR$\uparrow$ & CR\%$\uparrow$ & MDD\%$\downarrow$ & WR\%$\uparrow$ & Vol\%$\downarrow$
& SR$\uparrow$ & CR\%$\uparrow$ & MDD\%$\downarrow$ & WR\%$\uparrow$ & Vol\%$\downarrow$ \\
\midrule
\rowcolor{rulemethodrow}
\multicolumn{16}{c}{\textit{\textbf{Rule-Based Baselines}}} \\
SPY
& 2.77 & 2.94 & 2.50 & 63.2 & 0.90
& -0.02 & -1.15 & 12.05 & 60.0 & 3.41
& 8.22 & 4.34 & 1.32 & 65.0 & 0.41 \\
B\&H
& 4.21 & 4.11 & 2.06 & 52.6 & 0.81
& -0.44 & -2.48 & 10.90 & 50.0 & 3.01
& 8.35 & 5.79 & 0.82 & 75.0 & 0.54 \\
MACD
& 3.16 & 1.05 & 1.02 & 52.6 & 0.28
& -1.45 & -1.37 & 4.38 & 35.0 & 0.73
& 5.72 & 2.29 & 0.58 & 65.0 & 0.32 \\
KDJ\&RSI
& 4.98 & 1.34 & 0.54 & 57.9 & 0.22
& 2.22 & 5.13 & 4.33 & 50.0 & 1.91
& 3.54 & 0.38 & 0.21 & 40.0 & 0.09 \\
ZMR
& 5.62 & 3.70 & 0.58 & 63.2 & 0.55
& 1.23 & 3.49 & 7.82 & 55.0 & 2.64
& 5.82 & 1.13 & 0.30 & 55.0 & 0.15 \\
SMA
& 5.14 & 3.27 & 1.46 & 68.4 & 0.53
& -3.99 & -2.43 & 2.67 & 40.0 & 0.49
& 9.58 & 5.31 & 0.43 & 75.0 & 0.43 \\
\midrule
\rowcolor{llmmethodrow}
\multicolumn{16}{c}{\textit{\textbf{LLM-Based Methods (Model: GPT-5-mini)}}} \\
Static Base Agent
& 1.55 & 1.89 & 2.82 & 47.4 & 1.07
& \underline{-1.66} & \underline{-5.06} & 9.86 & 41.7 & 2.26
& 5.88 & 4.15 & 1.06 & 66.7 & 0.55 \\
Static TC Agent
& 2.87 & 3.30 & \underline{2.79} & \underline{50.9} & \underline{0.97}
& \textbf{-1.54} & \textbf{-4.71} & 9.77 & \textbf{51.7} & 2.23
& 6.45 & \underline{4.86} & 1.37 & \underline{78.3} & 0.59 \\
EvolveBase
& 1.41 & 1.82 & 3.17 & 46.7 & 1.06
& -2.87 & -7.58 & 10.30 & 44.4 & \underline{2.01}
& 5.29 & 3.80 & \underline{1.04} & 69.8 & \underline{0.54} \\
EvolveStrategy
& \underline{3.86} & \underline{4.70} & 3.18 & 49.1 & 1.01
& -3.09 & -6.85 & \textbf{8.07} & 45.0 & \textbf{1.74}
& \underline{7.76} & 4.78 & \textbf{0.65} & 70.0 & \textbf{0.48} \\
EvolveTrade (Ours)
& \textbf{5.12} & \textbf{5.10} & \textbf{2.39} & \textbf{57.0} & \textbf{0.84}
& -2.53 & -6.60 & \underline{9.13} & \underline{46.7} & \underline{2.01}
& \textbf{8.43} & \textbf{6.84} & 1.26 & \textbf{81.3} & 0.65 \\

\bottomrule
\end{tabular}
}%
\end{table*}

\begin{table*}[t]
\centering
\small

\setlength{\tabcolsep}{3.0pt}
\resizebox{\textwidth}{!}{%
\begin{tabular}{lccccccccccccccc}
\toprule
& \multicolumn{5}{c}{\textbf{Nov. 2025}}
& \multicolumn{5}{c}{\textbf{Feb. 2026}}
& \multicolumn{5}{c}{\textbf{Apr. 2026}} \\
\cmidrule(lr){2-6}
\cmidrule(lr){7-11}
\cmidrule(lr){12-16}
\textbf{Methods}
& SR$\uparrow$ & CR\%$\uparrow$ & MDD\%$\downarrow$ & WR\%$\uparrow$ & Vol\%$\downarrow$
& SR$\uparrow$ & CR\%$\uparrow$ & MDD\%$\downarrow$ & WR\%$\uparrow$ & Vol\%$\downarrow$
& SR$\uparrow$ & CR\%$\uparrow$ & MDD\%$\downarrow$ & WR\%$\uparrow$ & Vol\%$\downarrow$ \\
\midrule
\rowcolor{rulemethodrow}
\multicolumn{16}{c}{\textit{\textbf{Rule-Based Baselines}}} \\
SPY
& 0.08 & 0.01 & 4.51 & 61.1 & 0.99
& -1.34 & -1.35 & 2.56 & 44.4 & 0.86
& 9.80 & 9.68 & 0.85 & 70.0 & 0.75 \\
B\&H
& 0.16 & 0.09 & 3.68 & 55.6 & 0.79
& 0.33 & 0.23 & 1.80 & 55.6 & 0.75
& 11.30 & 9.23 & 0.45 & 65.0 & 0.62 \\
MACD
& 3.99 & 0.92 & 0.34 & 55.6 & 0.20
& -5.82 & -1.80 & 2.38 & 33.3 & 0.27
& 10.63 & 7.39 & 0.53 & 70.0 & 0.54 \\
KDJ\&RSI
& 3.63 & 1.39 & 1.38 & 55.6 & 0.34
& 0.68 & 0.23 & 0.69 & 50.0 & 0.30
& 5.10 & 0.98 & 0.32 & 60.0 & 0.15 \\
ZMR
& 2.80 & 1.15 & 1.39 & 55.6 & 0.37
& -1.49 & -0.96 & 1.92 & 50.0 & 0.56
& 11.35 & 4.10 & 0.21 & 85.0 & 0.28 \\
SMA
& -1.04 & -0.70 & 2.62 & 61.1 & 0.57
& 2.26 & 1.09 & 0.78 & 55.6 & 0.43
& 5.72 & 1.37 & 0.52 & 60.0 & 0.19 \\

\midrule
\rowcolor{llmmethodrow}
\multicolumn{16}{c}{\textit{\textbf{LLM-Based Methods (Model: Gemini-2.5-Flash)}}} \\
Static Base Agent
& -1.89 & -1.73 & 4.24 & 46.3 & \underline{0.79}
& -2.04 & -1.83 & 3.03 & 40.7 & 0.78
& \textbf{9.07} & \textbf{11.64} & 1.27 & \textbf{78.3} & 0.97 \\
Static TC Agent
& -1.99 & -2.24 & 5.22 & 50.0 & 0.97
& 2.70 & 2.29 & \underline{2.10} & 48.1 & \underline{0.75}
& 6.62 & 4.73 & \textbf{0.78} & 60.0 & \underline{0.55} \\
EvolveBase
& -2.28 & -2.16 & \underline{4.07} & 49.1 & \textbf{0.78}
& -1.70 & -1.38 & 3.19 & 40.4 & \textbf{0.67}
& \underline{7.98} & \underline{10.50} & 1.40 & \underline{69.8} & 0.96 \\
EvolveStrategy
& \underline{-1.04} & \underline{-1.16} & \textbf{4.00} & \underline{51.9} & 0.91
& \textbf{3.92} & \textbf{3.52} & \textbf{2.02} & \textbf{51.9} & 0.78
& 3.66 & 2.23 & 1.09 & 50.0 & \textbf{0.49} \\
EvolveTrade (Ours)
& \textbf{-1.03} & \textbf{-1.07} & 4.49 & \textbf{53.7} & 0.86
& \underline{2.75} & \underline{2.92} & 2.66 & \textbf{51.9} & 0.95
& 4.73 & 3.69 & \underline{0.94} & 55.0 & 0.59 \\

\bottomrule
\end{tabular}
}%
\caption{Main results across backbone models in market regimes after each model's knowledge cutoff. Each panel corresponds to one backbone model, and LLM-based rows report averages over three runs. Bold and underline mark the best and second-best LLM-based methods, respectively, for each metric within each model-regime setting.}
\label{tab:main_result}
\end{table*}

\subsection{Online Policy Self-Evolution}
\label{sec:method:refinement}

EvolveTrade applies refinement steps at a fixed period of $N \geq 1$ trading days, which controls how frequently the policy is revised.
We partition the trading horizon into consecutive batches of $N$ days, so that batch $k$ covers days $(k{-}1)N+1, \ldots, kN$.
Within a batch the policy is held fixed: $\pi_{(k-1)N+1} = \cdots = \pi_{kN}$.
At the end of batch $k$, EvolveTrade collects records of the batch:
\begin{equation}
    \mathcal{B}_k = \{\mathcal{R}_{(k-1)N+1}, \ldots, \mathcal{R}_{kN}\},
\end{equation}
and a separate \emph{Policy Agent} applies a language-based refinement step to obtain the policy used in batch $k+1$:
\begin{equation}
    \pi_{kN+1} = f_{\mathrm{update}}(\pi_{kN}, \mathcal{B}_k).
\end{equation}

The Policy Agent is implemented as an LLM call with a fixed update instruction and textual inputs:
\begin{equation}
    f_{\mathrm{update}}(\pi, \mathcal{B}) = \mathrm{LLM}\big(I_{\mathrm{update}}, \pi, \mathcal{B}\big).
\end{equation}
Here, $I_{\mathrm{update}}$ asks the Policy Agent to analyze the records in $\mathcal{B}$, identify which aspects of the current prompt most contributed to good or poor performance, and rewrite the full policy text accordingly (\autoref{fig:prompt-evolvetrade-update}).
The instruction emphasizes grounded edits over generic rewrites and asks the agent to link concrete observations in the records to specific lines of the current prompt.

A refinement step may revise how the agent selects tools, formulates queries, interprets signals, verifies evidence, controls risk, or structures its final output.
For example, if the trajectories in $\mathcal{B}_k$ show that the agent repeatedly acted on news signals without checking price confirmation, the update may introduce a rule requiring cross-validation before increasing exposure.

Iterating the refinement step across the trading horizon yields a sequence of batch policies in which the policy active during batch $k+1$ has been shaped by the agent's own realized experience over batches $1, \ldots, k$.
The policy active on a later trading day is therefore a self-evolved policy rather than the initial prompt the agent started with.

Treating the policy text as an editable variable updated from feedback aligns with the broader view that natural-language prompts can be optimized in compound LLM systems~\citep{textgrad}.
The realized performance of the resulting policy sequence under repeated online refinement is what we evaluate in \autoref{sec:experiments}.

\section{Experiment}
\label{sec:experiments}
\subsection{Experiment Setup}

\paragraph{Trading Environment.}
We use a daily-close simulation environment based on the LiveTradeBench~\cite{livetradebench} framework.
On each trading day, the backtesting system tracks the current portfolio, receives target percentage weights from the agent, and rebalances the portfolio at the daily closing price.
The investment universe consists of 15 major US blue-chip stocks (e.g., \texttt{AAPL}, \texttt{MSFT}, \texttt{NVDA}, \texttt{JPM}) and a cash component.

\paragraph{Baselines.}
We evaluate rule-based trading baselines and three LLM-based configurations.
\textbf{Rule-based baselines} use fixed trading rules; they include \textbf{SPY}, \textbf{Buy and Hold (B\&H)}, \textbf{MACD}, \textbf{KDJ\&RSI}, \textbf{ZMR}, and \textbf{SMA}.
\autoref{sec:appendix:rule-baselines} provides the definitions and allocation method used for these rule-based baselines.
\textbf{LLM-based configurations} isolate the effect of tool access and policy refinement.
\textbf{Static Base Agent} follows the basic \texttt{Live-Trade-Bench} design~\citep{livetradebench}, using a fixed daily observation window of price and news context without using tools.
\textbf{Static Tool-Calling Agent} uses the same trading tools as EvolveTrade but keeps its policy fixed throughout the evaluation.
\textbf{EvolveBase} follows the same tool-free design as the Static Base Agent but allows its policy to self-evolve, isolating the effect of policy evolution in the absence of tool access.
\textbf{EvolveStrategy} utilizes a Policy Agent to adaptively update the Trading Agent's high-level trading strategies, while keeping its operational and tool-calling protocols fixed. The core framework of its Policy Agent is adapted from the design of ATLAS~\citep{atlas}, inheriting its approach to strategy-level instruction evolution.
\textbf{EvolveTrade (Ours)} employs a Policy Agent that comprehensively updates the Trading Agent's entire policy, specifically focusing on adapting tool-calling protocols.

\paragraph{Implementation Details.}
We evaluate the LLM-based configurations using two backbone models, \texttt{GPT-5-mini}~\citep{gpt5} and \texttt{Gemini-2.5-Flash}~\citep{gemini25flash}, with the Trading Agent and Policy Agent sharing a backbone within each run and policy updates applied every $N=5$ trading days; initial policy templates and update prompts are in \appref{app:prompts}.

\paragraph{Evaluation.}
We evaluate each agent over one-month online policy-updating windows, following the live evaluation length of DeepFund~\citep{deepfund}, in which the agent repeatedly observes market evidence, submits a daily allocation, receives portfolio feedback, and (for EvolveTrade) updates its policy at the specified interval.
To test whether policy evolution generalizes across market regimes, \texttt{GPT-5-mini} is evaluated on \textbf{January, April, and September 2025} (sideways, drawdown-recovery, and uptrend regimes), and \texttt{Gemini-2.5-Flash} on \textbf{November 2025, February 2026, and April 2026} (bearish, sideways, and bullish regimes) to account for its later data cutoff.

We report annualized Sharpe Ratio (SR), cumulative return (CR), maximum drawdown (MDD), win rate (WR), and daily volatility (Vol), with definitions in \autoref{sec:appendix:evaluation-metrics}; each LLM-based configuration is run three times, and tables report averages.

\begin{figure*}[t]
    \centering
    \includegraphics[width=\textwidth]{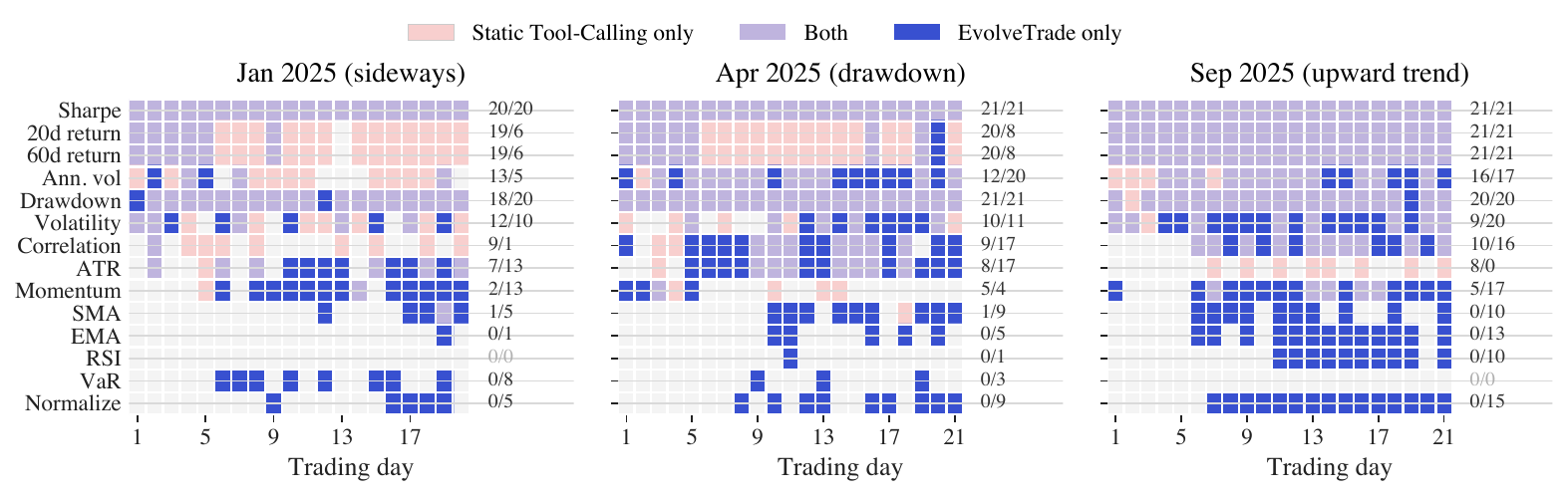}
    \vspace{-0.3in}
    \caption{
    \textbf{EvolveTrade activates regime-relevant analytic metrics that the Static Tool-Calling baseline never invokes.}
    Per-day binary map of analytic metrics appearing in each agent's Python code interpreter calls.
    \emph{red} cells mark days where only the Static Tool-Calling baseline invokes the metric, \emph{purple} cells mark days where both agents invoke it, and \emph{blue} cells mark days where only EvolveTrade invokes it.
    Right-margin counts are formatted as \texttt{Static Tool-Calling days / EvolveTrade days} over the corresponding regime.}
    \vspace{-0.2in}
    \label{fig:coverage-unlocked}
\end{figure*}

\subsection{Main Result}

\paragraph{EvolveTrade improves LLM trading agents.}
\autoref{tab:main_result} compare the agents across market regimes after each model's knowledge cutoff (standard deviations across the three runs are reported in \autoref{sec:appendix:main-results-std}).
Across the LLM-based methods, EvolveTrade attains the best result for many metric-regime pairs, indicating that updating the tool-use policy generally improves the agent's risk-return profile beyond both the Static Agent and the Static TC Agent.
For \texttt{GPT-5-mini}, EvolveTrade is strongest in the January and September periods, achieving the best SR and CR among LLM agents while also improving several risk-related metrics.
For \texttt{Gemini-2.5-Flash}, EvolveTrade performs best in the November and second-best in the February, suggesting that the benefit of policy evolution transfers beyond a single backbone model.

Notably, these gains are not limited to comparisons against LLM baselines.
In some regimes, EvolveTrade also exceeds all rule-based baselines on return-side metrics.
For example, with \texttt{GPT-5-mini} in September 2025, EvolveTrade reaches a CR of 6.84\%, outperforming the strongest rule-based CR in that period.
Similarly, under Gemini-2.5-Flash in February 2026, EvolveTrade achieves an SR of 2.75 and a CR of 2.92\%, consistently outperforming all traditional rule-based baselines. 
By mitigating LLM vulnerabilities in non-stationary environments through continuous tool-use policy updates, EvolveTrade delivers the most dependable risk-adjusted performance among all LLM-based methods across shifting market conditions.

\begin{figure}[t]
    \centering
    \includegraphics[width=\columnwidth]{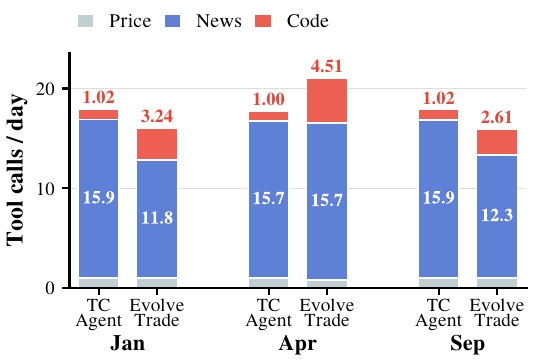}
    \caption{Tool-use behavior under the TC Agent (Static Tool-Calling Agent) and EvolveTrade for \texttt{GPT-5-mini}. Bars report mean daily tool calls over three runs.}
    \label{fig:tool-use-behavior}
    \vspace{-0.1in}
\end{figure}

\begin{figure}[t]
    \centering
    \includegraphics[width=\columnwidth]{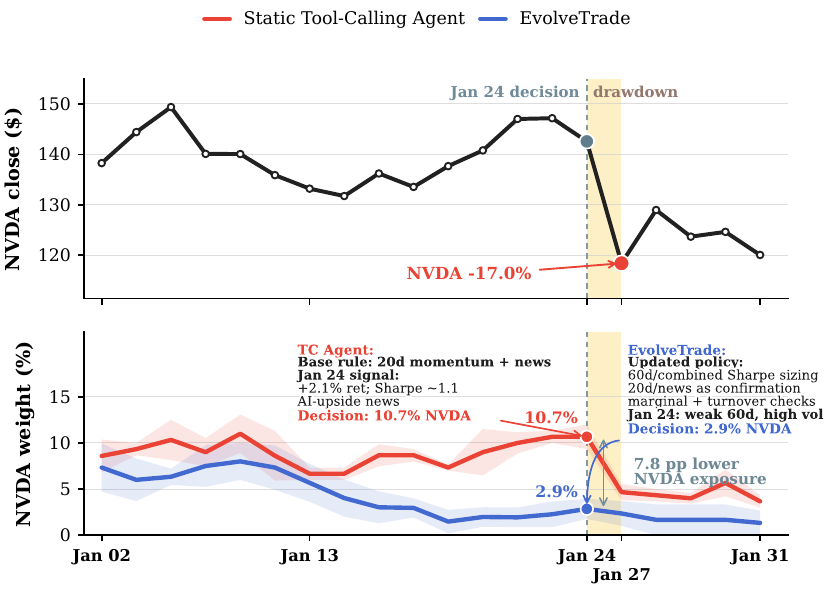}
    \caption{Case study around the January 2025 NVDA drawdown. On January 24, the Static Tool-Calling Agent held 10.7\% NVDA, whereas EvolveTrade held 2.9\% under its refined sizing policy. NVDA fell 17.0\% on the next trading day.}
    \vspace{-0.1in}
    \label{fig:nvda-case-study}
\end{figure}

\subsection{Analysis}
\paragraph{EvolveTrade activates regime-relevant metrics the baseline never invokes.}
\autoref{fig:coverage-unlocked} reports, for each trading day in each regime, which analytic metrics appear in the agent's code-interpreter calls under Static Tool-Calling only (red), under both agents (purple), or under EvolveTrade only (blue).
The Static Tool-Calling baseline never invokes EMA, RSI, VaR, or signal normalization in any of the three regimes, while EvolveTrade activates these regime-relevant metrics once policy self-evolution begins: tail-risk metrics (VaR, signal normalization) in the April drawdown, and trend-following indicators (SMA, EMA, RSI) in the September upward trend.
The blue cells in the bottom rows show this expansion of analytic vocabulary across all three regimes, while the red cells in the upper rows show that EvolveTrade also drops some of the baseline's repetitive metric calls when they are no longer informative.

\paragraph{EvolveTrade changes tool-use patterns.}
\autoref{fig:tool-use-behavior} compares the daily tool-call frequency of the TC Agent and EvolveTrade.
The price-retrieval tool ($t_{\text{price}}$) remains close to one call per day in both settings, while the code-execution tool ($t_{\text{code}}$) shows a clear shift under EvolveTrade, rising from about 1.0 to 2.6--4.5 calls per day across regimes; the news-search tool ($t_{\text{news}}$) does not increase overall.
Together with the metric-level evidence in \autoref{fig:coverage-unlocked}, this suggests policy refinement redirects tool use from external text gathering toward executable analysis, changing the agent's tool-use strategy rather than merely its access to tools.

\paragraph{EvolveTrade limits NVDA exposure before the drawdown.}
\autoref{fig:nvda-case-study} illustrates how policy refinement changes a concrete allocation.
On January 24, the Static Tool-Calling Agent holds 10.7\% NVDA on average, relying on a +2.1\% 20-day return, a 20-day Sharpe around 1.1, and positive AI news.
In contrast, EvolveTrade holds only 2.9\% under a refined policy that prioritizes longer-horizon risk-adjusted evidence (60-day or combined Sharpe), treats short-horizon metrics and news as confirmation, and applies portfolio-improvement and turnover checks when sizing volatile positions.
NVDA falls 17.0\% on the next trading day, and EvolveTrade returns -0.03\% compared with -1.11\% for the Static Tool-Calling Agent.
This case provides a traceable link between policy refinement, portfolio exposure, and realized return; a complementary example is provided in \autoref{sec:appendix:evolvetrade-trace-case}.

\begin{table}[t]
\centering
\small

\setlength{\tabcolsep}{4.0pt}
\resizebox{\columnwidth}{!}{%
\begin{tabular}{lccccc}
\toprule
\textbf{Method}
& SR$\uparrow$ & CR\%$\uparrow$ & MDD\%$\downarrow$ & WR\%$\uparrow$ & Vol\%$\downarrow$ \\
\midrule
Static Base Agent      & 1.82 & 4.51  & 4.70 & \textbf{68.0} & 0.82 \\
Static TC Agent   & 2.94 & 8.88  & 4.40 & 63.3 & 0.95 \\
EvolveBase        & 2.41 & 4.79  & 3.00 & 65.3 & \textbf{0.65} \\
EvolveStrategy    & 3.27 & 6.94  & \textbf{2.82} & 65.3 & 0.68 \\
EvolveTrade (Ours) & \textbf{4.00} & \textbf{10.56} & 2.96 & 67.3 & 0.81 \\
\bottomrule
\end{tabular}%
}
\caption{Performance comparison over 50 trading days using GPT-5-mini (2025-09-02 -- 2025-11-10).}
\label{tab:long_horizon_gpt}
\end{table}

\begin{figure}[t]
    \centering
    \includegraphics[width=0.8\columnwidth] {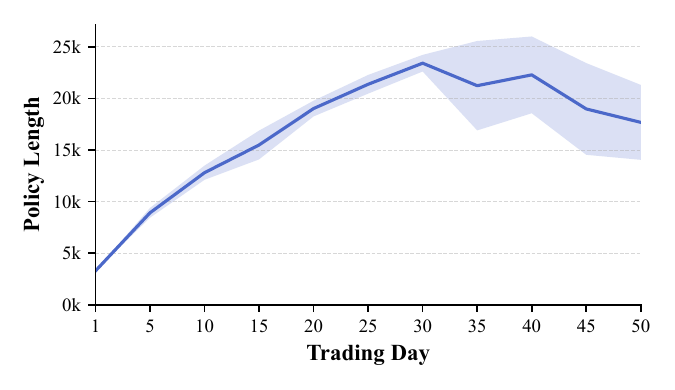}
    \vspace{-0.1in}
    \caption{Policy length under EvolveTrade over 50 trading days across 3 runs. (2025-09-02 -- 2025-11-10)}
    \label{fig:policy-prompt-growth-50-trading-day}
    \vspace{-0.1in}
\end{figure}

\paragraph{EvolveTrade remains effective over a longer trading horizon.}
Repeated policy updates can lengthen the policy and potentially impair the agent over extended trading horizons.
We therefore evaluate EvolveTrade with \texttt{GPT-5-mini} over 50 trading days.
As shown in \autoref{tab:long_horizon_gpt}, EvolveTrade outperforms all baselines in both SR and CR, indicating that its effectiveness persists under continued online refinement.
During the same evaluation, \autoref{fig:policy-prompt-growth-50-trading-day} tracks policy length across three independent runs.
The policy grows as new guidance is incorporated but contracts periodically, a pattern consistent with consolidating redundant instructions rather than accumulating them indefinitely.
Together, the sustained performance and non-monotonic policy growth suggest that EvolveTrade can continue adapting over longer horizons without requiring an ever-expanding policy.

\begin{table}[t]
\centering
\small
\setlength{\tabcolsep}{4.5pt}
\resizebox{\columnwidth}{!}{%
\begin{tabular}{lccccc}
\toprule
\textbf{Update Interval} & \textbf{SR}$\uparrow$ & \textbf{CR\%}$\uparrow$ & \textbf{MDD\%}$\downarrow$ & \textbf{WR\%}$\uparrow$ & \textbf{Vol\%}$\downarrow$ \\
\midrule
1 day & 1.92 & 0.16 & 4.58 & 61.5 & 1.13 \\
3 days & 3.25 & 0.97 & 4.46 & \textbf{62.7} & \textbf{1.13} \\
5 days & \textbf{3.67} & \textbf{1.78} & \textbf{4.26} & 61.7 & 1.17 \\
7 days & 3.50 & 1.54 & 4.55 & 60.6 & 1.21 \\
\bottomrule
\end{tabular}
}
\caption{Sensitivity to policy-update interval. Results are averaged over January, April, and September 2025.}
\label{tab:policy_update_interval}
\vspace{-0.15in}
\end{table}

\paragraph{Moderate update intervals balance adaptation and stability.}
Varying the Policy Agent's revision frequency (\autoref{tab:policy_update_interval}) reveals a non-monotonic relationship: the best-performing interval achieves the highest average SR and CR across January, April, and September 2025, while a shorter interval yields the highest WR and a comparable lowest Vol.
In contrast, daily updates produce the lowest average SR and CR, suggesting overly frequent revision overfits short-horizon feedback or amplifies noise from individual trading days, and that policy evolution benefits from enough feedback accumulation between updates.

\section{Conclusion}

We presented \textbf{EvolveTrade}, a self-evolving framework that treats the system prompt of a tool-using LLM trading agent as a text-parameterized policy for evidence gathering, tool use, risk management, and portfolio allocation.
EvolveTrade periodically revises this policy from the agent's tool-use trajectories and realized portfolio feedback, allowing its procedure to adapt as market conditions unfold.
Across multiple market regimes and two LLM backbones, EvolveTrade improves risk-adjusted performance over static baselines and changes tool-use behavior, including more code-mediated signal checks and a case-level policy-to-return attribution linking one allocation change to realized performance.
These results suggest that robust LLM trading agents must continually understand changing markets and refine the policies by which they act on evidence.

\section*{Limitations}

Our sensitivity analysis evaluates several policy-update intervals and finds that moderate intervals provide a stable balance between adaptation and performance.
The current implementation nevertheless keeps the selected interval fixed throughout each run, leaving dynamically scheduled policy updates in response to market changes as a promising direction for future work.
We also evaluate portfolio turnover and performance under proportional transaction costs in \autoref{sec:appendix:trading-friction}, but these estimates do not capture slippage, market impact, or liquidity constraints.
Future evaluation could incorporate these factors through more realistic execution models.


\bibliography{custom}

\clearpage
\appendix

\begin{table*}[t!]
\centering
\small
\setlength{\tabcolsep}{5.0pt}
\begin{tabular}{lcccccc}
\toprule
& \multicolumn{2}{c}{\textbf{Jan. 2025}}
& \multicolumn{2}{c}{\textbf{Apr. 2025}}
& \multicolumn{2}{c}{\textbf{Sep. 2025}} \\
\cmidrule(lr){2-3}
\cmidrule(lr){4-5}
\cmidrule(lr){6-7}
\textbf{Methods}
& SR$\uparrow$ & CR\%$\uparrow$
& SR$\uparrow$ & CR\%$\uparrow$
& SR$\uparrow$ & CR\%$\uparrow$ \\
\midrule
Static Base Agent
& 1.55$\pm$0.14 & 1.89$\pm$0.15
& \underline{-1.66}$\pm$0.25 & \underline{-5.06}$\pm$0.66
& 5.88$\pm$0.63 & 4.15$\pm$0.52 \\
Static TC Agent
& 2.87$\pm$0.19 & 3.30$\pm$0.17
& \textbf{-1.54}$\pm$0.11 & \textbf{-4.71}$\pm$0.36
& 6.45$\pm$0.87 & \underline{4.86}$\pm$0.56 \\
EvolveBase
& 1.41$\pm$0.40 & 1.82$\pm$0.64
& -2.87$\pm$0.75 & -7.58$\pm$0.77
& 5.29$\pm$1.02 & 3.80$\pm$0.76 \\
EvolveStrategy
& \underline{3.86}$\pm$1.51 & \underline{4.70}$\pm$1.84
& -3.09$\pm$1.42 & -6.85$\pm$3.13
& \underline{7.76}$\pm$0.47 & 4.78$\pm$1.07 \\
EvolveTrade (Ours)
& \textbf{5.12}$\pm$0.51 & \textbf{5.10}$\pm$0.72
& -2.53$\pm$0.47 & -6.60$\pm$1.19
& \textbf{8.43}$\pm$1.68 & \textbf{6.84}$\pm$0.70 \\
\bottomrule
\end{tabular}
\caption{Main results using \texttt{GPT-5-mini} with standard deviation across three runs (mean$\pm$std). Bold and underline mark the best and second-best LLM-based methods, respectively.}
\label{tab:main_results_std_gpt}
\end{table*}

\begin{table*}[t]
\centering
\small
\setlength{\tabcolsep}{5.0pt}
\begin{tabular}{lcccccc}
\toprule
& \multicolumn{2}{c}{\textbf{Nov. 2025}}
& \multicolumn{2}{c}{\textbf{Feb. 2026}}
& \multicolumn{2}{c}{\textbf{Apr. 2026}} \\
\cmidrule(lr){2-3}
\cmidrule(lr){4-5}
\cmidrule(lr){6-7}
\textbf{Methods}
& SR$\uparrow$ & CR\%$\uparrow$
& SR$\uparrow$ & CR\%$\uparrow$
& SR$\uparrow$ & CR\%$\uparrow$ \\
\midrule
Static Base Agent
& -1.89$\pm$0.16 & -1.73$\pm$0.19
& -2.04$\pm$0.26 & -1.83$\pm$0.21
& \textbf{9.07}$\pm$0.60 & \textbf{11.64}$\pm$0.96 \\
Static TC Agent
& -1.99$\pm$0.18 & -2.24$\pm$0.10
& 2.70$\pm$0.23 & 2.29$\pm$0.25
& 6.62$\pm$0.78 & 4.73$\pm$1.11 \\
EvolveBase
& -2.28$\pm$0.23 & -2.16$\pm$0.26
& -1.70$\pm$0.68 & -1.38$\pm$0.50
& \underline{7.98}$\pm$0.59 & \underline{10.50}$\pm$0.22 \\
EvolveStrategy
& \underline{-1.04}$\pm$0.47 & \underline{-1.16}$\pm$0.52
& \textbf{3.92}$\pm$1.03 & \textbf{3.52}$\pm$1.13
& 3.66$\pm$1.49 & 2.23$\pm$0.84 \\
EvolveTrade (Ours)
& \textbf{-1.03}$\pm$0.39 & \textbf{-1.07}$\pm$0.40
& \underline{2.75}$\pm$0.29 & \underline{2.92}$\pm$0.25
& 4.73$\pm$1.48 & 3.69$\pm$1.83 \\
\bottomrule
\end{tabular}
\caption{Main results using \texttt{Gemini-2.5-Flash} with standard deviation across three runs (mean$\pm$std). Bold and underline mark the best and second-best LLM-based methods, respectively.}
\label{tab:main_results_std_gemini}
\end{table*}

\section{Rule-Based Baseline Definitions}
\label{sec:appendix:rule-baselines}

We include standard market and technical-analysis baselines commonly used in LLM trading-agent evaluations, following the baseline family used by TradingAgents~\citep{tradingagents} while adapting the definitions to our daily-close portfolio setting.
All rule-based baselines are deterministic, use only historical price information available up to the decision date, and do not use news, LLM reasoning, tool selection, or policy refinement.

\textbf{SPY} is a market benchmark that buys and holds the SPY index ETF over the evaluation window.
\textbf{B\&H} initializes an equal-weight portfolio over the tradable equities at the start of the evaluation window and holds the positions without signal-based reallocation.
\textbf{MACD} is a momentum rule based on the moving-average convergence divergence indicator: it increases exposure after bullish MACD--signal-line crossovers and reduces exposure after bearish crossovers.
\textbf{KDJ\&RSI} combines stochastic-oscillator information from KDJ with the relative strength index; it treats oversold reversals as buy signals and overbought or weakening momentum states as sell or risk-reduction signals.
\textbf{ZMR} is a mean-reversion rule that reacts to normalized price deviations from a recent reference level, increasing exposure when prices are sufficiently below the reference and reducing exposure after reversion or positive deviation.
\textbf{SMA} is a moving-average crossover rule that takes risk-on positions when a short-horizon simple moving average is above a longer-horizon moving average and moves risk-off when the short-horizon average falls below the long-horizon average.

To ensure uniform exposure and prevent single-asset concentration, all active strategies enforce a strict per-asset position cap of $1/N$ (where $N=15$, representing the 15 tradable equities in our universe). When a \textit{buy} signal is generated for a specific asset, the strategy enters a risk-on position by allocating an equal weight of $1/15$ ($\approx 6.7\%$) to that asset. Conversely, when a \textit{sell} signal occurs, the strategy completely exits the position, reducing the asset's exposure to zero. Any unallocated capital remains in cash earning zero return.

\section{Evaluation Metric Definitions}
\label{sec:appendix:evaluation-metrics}

Let $V_t$ denote the portfolio value after the close of trading day $t$, and let the daily portfolio return be
\begin{equation}
    R_t = \frac{V_t - V_{t-1}}{V_{t-1}}.
\end{equation}
The main tables report the following metrics.
These metrics follow common trading-agent evaluation practice, including LiveTradeBench~\citep{livetradebench} and TradingAgents~\citep{tradingagents}.
\textbf{Sharpe Ratio (SR)} measures risk-adjusted performance:
\begin{equation}
    \mathrm{SR} =
    \sqrt{252}
    \frac{\bar{R}}{\sigma_R},
    \quad
    \sigma_R =
    \sqrt{
        \frac{1}{T-1}
        \sum_{t=1}^{T}
        (R_t - \bar{R})^2
    },
\end{equation}
where $\bar{R}$ is the mean daily return and $\sigma_R$ is the sample standard deviation of daily returns.
We omit a risk-free-rate adjustment because the evaluation windows are one month long and all methods are compared on the same periods.
\textbf{Cumulative return (CR)} measures total portfolio growth over the evaluation window:
\begin{equation}
    \mathrm{CR} = \left(\frac{V_T}{V_0} - 1\right) \times 100.
\end{equation}
\textbf{Maximum drawdown (MDD)} measures the largest percentage decline from a previous portfolio peak:
\begin{equation}
    \mathrm{MDD} =
    \max_{1 \leq t \leq T}
    \left(
        \frac{\max_{0 \leq s \leq t} V_s - V_t}{\max_{0 \leq s \leq t} V_s}
    \right)
    \times 100.
\end{equation}
\textbf{Win rate (WR)} is the fraction of trading days with positive portfolio return:
\begin{equation}
    \mathrm{WR} =
    \frac{1}{T}
    \sum_{t=1}^{T}
    \mathbb{I}[R_t > 0]
    \times 100.
\end{equation}
\textbf{Daily volatility (Vol)} is the sample standard deviation of daily returns, reported in percentage points:
\begin{equation}
    \mathrm{Vol} =
    \sqrt{
        \frac{1}{T-1}
        \sum_{t=1}^{T}
        (R_t - \bar{R})^2
    }
    \times 100.
\end{equation}

\section{Main Results with Standard Deviation}
\label{sec:appendix:main-results-std}
To assess run-to-run reliability, we repeat each LLM-agent experiment three times and report the mean and standard deviation of SR and CR in \autoref{tab:main_results_std_gpt} and \autoref{tab:main_results_std_gemini}.
EvolveTrade preserves the main performance pattern across repeated runs, achieving the strongest average SR and CR among the LLM-based methods in January and September with \texttt{GPT-5-mini}, and in November with \texttt{Gemini-2.5-Flash}.
Moreover, its standard deviations are generally comparable to those of the other policy-evolving agents, with no systematic increase in run-to-run variability.
These results show that EvolveTrade's gains persist across independent policy-evolution trajectories rather than arising from a single favorable run.
\begin{table*}[t]
\centering
\small
\setlength{\tabcolsep}{2.5pt}
\resizebox{\textwidth}{!}{%
\begin{tabular}{lcccccccccccccccccc}
\toprule
& \multicolumn{9}{c}{\textit{\textbf{GPT-5-mini}}}
& \multicolumn{9}{c}{\textit{\textbf{Gemini-2.5-Flash}}} \\
\cmidrule(lr){2-10}
\cmidrule(lr){11-19}
& \multicolumn{3}{c}{\textbf{Jan. 2025}}
& \multicolumn{3}{c}{\textbf{Apr. 2025}}
& \multicolumn{3}{c}{\textbf{Sep. 2025}}
& \multicolumn{3}{c}{\textbf{Nov. 2025}}
& \multicolumn{3}{c}{\textbf{Feb. 2026}}
& \multicolumn{3}{c}{\textbf{Apr. 2026}} \\
\cmidrule(lr){2-4}
\cmidrule(lr){5-7}
\cmidrule(lr){8-10}
\cmidrule(lr){11-13}
\cmidrule(lr){14-16}
\cmidrule(lr){17-19}
\textbf{Methods}
& TO$\downarrow$ & SR$\uparrow$ & CR\%$\uparrow$
& TO$\downarrow$ & SR$\uparrow$ & CR\%$\uparrow$
& TO$\downarrow$ & SR$\uparrow$ & CR\%$\uparrow$
& TO$\downarrow$ & SR$\uparrow$ & CR\%$\uparrow$
& TO$\downarrow$ & SR$\uparrow$ & CR\%$\uparrow$
& TO$\downarrow$ & SR$\uparrow$ & CR\%$\uparrow$ \\
\midrule
Static Base
& 1.08 & 1.47 & 1.78
& 1.72 & \underline{-1.72} & \underline{-5.23}
& \underline{1.09} & 5.72 & 4.04
& 2.71 & -2.19 & -2.00
& 2.48 & -2.32 & -2.08
& 2.43 & \textbf{8.86} & \textbf{11.37} \\
Static TC
& 2.25 & 2.68 & 3.06
& 2.53 & \textbf{-1.63} & \textbf{-4.95}
& 1.90 & 6.19 & \underline{4.66}
& \underline{2.32} & -2.20 & -2.47
& \underline{2.27} & 2.44 & 2.06
& 3.21 & 6.15 & 4.39 \\
EvolveBase
& \textbf{0.70} & 1.39 & 1.74
& \textbf{1.08} & -2.98 & -7.68
& \textbf{0.56} & 5.35 & 3.74
& 2.91 & -2.65 & -2.44
& 2.94 & -2.11 & -1.67
& \underline{2.36} & \underline{8.01} & \underline{10.24} \\
EvolveStrategy
& 2.22 & \underline{3.66} & \underline{4.46}
& 3.32 & -3.24 & -7.16
& 2.98 & \underline{7.30} & 4.47
& 3.27 & \underline{-1.36} & \underline{-1.48}
& 3.04 & \textbf{3.58} & \textbf{3.21}
& 4.39 & 2.94 & 1.78 \\
EvolveTrade
& \underline{1.05} & \textbf{5.01} & \textbf{4.99}
& \underline{1.20} & -2.58 & -6.71
& 1.68 & \textbf{8.21} & \textbf{6.67}
& \textbf{1.93} & \textbf{-1.23} & \textbf{-1.26}
& \textbf{1.44} & \underline{2.62} & \underline{2.77}
& \textbf{1.90} & 4.48 & 3.49 \\
\bottomrule
\end{tabular}
}%
\caption{Turnover (TO) and performance after applying 10 bps transaction costs across backbone models and trading windows. Bold and underline mark the best and second-best LLM-based methods, respectively.}
\label{tab:trading_friction}
\end{table*}

\section{Main Results with Trading Friction}
\label{sec:appendix:trading-friction}
To better reflect the trading frictions encountered in real-world execution, we measure portfolio turnover and recompute SR and CR after applying a proportional transaction cost of 10 bps to traded portfolio value.
As shown in \autoref{tab:trading_friction}, EvolveTrade retains the highest SR and CR among the LLM-based methods in the same three settings where it leads in the main results: January and September with \texttt{GPT-5-mini}, and November with \texttt{Gemini-2.5-Flash}.
It also achieves the second-best SR and CR in February with \texttt{Gemini-2.5-Flash}, while maintaining lower turnover than both the Static Tool-Calling Agent and EvolveStrategy in all six evaluation windows.
The preserved rankings and consistently lower turnover show that EvolveTrade's strongest gains survive the transaction-cost adjustment rather than being explained by aggressive portfolio reallocation.

\section{Specific-Day Return Comparison for EvolveTrade}
\label{sec:appendix:evolvetrade-trace-case}

This appendix gives a matched single-day example from September 2025.
Both agents make a portfolio decision at the September 10 close, and the resulting portfolio is evaluated by the September 11 close.
The example is intended to show the intermediate mechanism behind the realized return gap, not to claim that the agent had access to September 11 information.
The Static Tool-Calling Agent uses code outputs as descriptive evidence before assigning qualitative sector weights.
EvolveTrade instead uses code as an allocation procedure: metric computation, signal calibration, target-weight construction, and validation.
This produces larger \texttt{JPM}, \texttt{CAT}, \texttt{WMT}, and \texttt{TSLA} exposures, which explain the next-day return advantage in \autoref{fig:appendix-evolvetrade-sep-trace}.

\section{Underperformance in the April Regimes}
\label{sec:appendix:april-underperformance}

EvolveTrade underperforms the strongest static baseline in both April windows: the Static Tool-Calling Agent for \texttt{GPT-5-mini} in April 2025 and the Static Base Agent for \texttt{Gemini-2.5-Flash} in April 2026. To characterize the allocation behavior associated with these results, we compare their average cash weights, retaining the same baseline across the other windows for each backbone.

\autoref{tab:appendix-april-cash} shows that EvolveTrade's average cash weight exceeds the selected baseline's by 10.1 and 35.9 percentage points in the two April windows, respectively, compared with at most 6.4 percentage points in the remaining windows. Despite their different market conditions, both April windows exhibit elevated cash allocation alongside weaker SR and CR, suggesting that reduced equity exposure may have contributed to missed gains during the April 2025 recovery and the April 2026 bullish period.




\begin{table}[t]
\centering
\small
\setlength{\tabcolsep}{4.0pt}
\begin{tabular}{lccc}
\toprule
\textbf{Period} & \textbf{Baseline CASH\%} & \textbf{ET CASH\%} & \textbf{$\Delta$cash} \\
\midrule
Jan. 2025 & 9.0 & 12.9 & +4.0 \\
\rowcolor{gray!15}
Apr. 2025 & 17.1 & 27.1 & +10.1 \\
Sep. 2025 & 8.7 & 8.0 & -0.7 \\
\midrule
Nov. 2025 & 9.7 & 16.1 & +6.4 \\
Feb. 2026 & 5.1 & 10.7 & +5.6 \\
\rowcolor{gray!15}
Apr. 2026 & 4.3 & 40.1 & +35.9 \\
\bottomrule
\end{tabular}
\caption{Average cash allocation of EvolveTrade (ET) versus Static TC for \texttt{GPT-5-mini} and Static Base for \texttt{Gemini-2.5-Flash}. $\Delta$cash is ET minus baseline in percentage points. April windows are highlighted.}
\label{tab:appendix-april-cash}
\end{table}

\section{Operational Prompts}
\label{app:prompts}
We provide the operational prompt templates used to instantiate EvolveTrade and EvolveStrategy.
The initial policy texts in \autoref{fig:prompt-evolvetrade-initial} and \autoref{fig:prompt-evolvestrategy-initial} guide how the Trading Agents operate: EvolveTrade specifies how the agent retrieves evidence, utilizes tools, and emits portfolio allocations, while EvolveStrategy specifies how the agent executes its trading decisions. Correspondingly, the policy-update prompts in \autoref{fig:prompt-evolvetrade-update} and \autoref{fig:prompt-evolvestrategy-update} outline how trajectory feedback is converted into the next iteration: EvolveTrade updates the entire policy framework including tool-calling protocols, whereas EvolveStrategy updates high-level trading strategies.
EvolveBase shares the same prompt structure as EvolveTrade, but is tool-free: price and news context are pre-fetched and inserted directly into the prompt, and its Fixed Block omits all tool-related instructions.

\section{Updated Policy Examples}
\label{app:updated-policy-examples}
This section presents the best-performing evolved policy for \texttt{EvolveTrade} under each backbone model, selected by the highest SR and CR during evaluation. The \texttt{GPT-5-mini} policy is shown in Figures~\ref{fig:evolved-policy-gpt-1}--\ref{fig:evolved-policy-gpt-3}, and the \texttt{Gemini-2.5-Flash} policy is shown in Figures~\ref{fig:evolved-policy-gemini-1}--\ref{fig:evolved-policy-gemini-2}; fixed structural blocks of the prompt template are excluded.

\clearpage
\begin{figure*}[t]
\centering
\begin{minipage}{\textwidth}
\small

\begin{tcolorbox}[targettrajstyle,title={Static Tool-Calling Agent}]
\textcolor{targettoolcall}{\textbf{[Policy: evidence collection only]}}\\
The fixed policy requires the agent to call \texttt{get\_price}, \texttt{code\_interpreter}, and \texttt{news\_searcher}, then write a one-line rationale for each asset.
The concrete instruction specifies an evidence format, not a sizing rule: each rationale should contain a price point, a code metric such as \texttt{20d return +3.18\%, Sharpe 0.91}, a news/context point, and a strategic rationale.
The policy does not define how much weight follows from that evidence, nor how metrics should be ranked, normalized, or converted into portfolio weights.\\[5pt]
\textcolor{targettoolcall}{\textbf{[Trajectory: metrics remain rationale text]}}\\
\textbf{Step 1.} \texttt{get\_price} retrieves all tickers from 2025-03-10 to 2025-09-10.\\
\textbf{Step 2.} \texttt{kernel\_code\_interpreter} makes one aggregate metric call for recent return, volatility, and Sharpe statistics.\\
\textbf{Step 3.} \texttt{news\_searcher} makes 15 per-asset calls, and the final allocation is chosen by qualitative synthesis.\\
\textbf{Observed trace pattern.} The same metrics are cited but do not force larger weights: \texttt{JPM} has 20d return +3.45\%, vol 27.11\%, Sharpe 2.105 and remains at 6\%; \texttt{CAT} has 20d return +2.23\%, vol 31.29\%, Sharpe 1.497 and remains at 4\%; \texttt{WMT} has Sharpe 1.227 and remains at 2\%; \texttt{TSLA} has Sharpe 1.625 but remains at 2\% because the rationale emphasizes high volatility.\\[5pt]
\textcolor{targettoolresult}{\textbf{[Portfolio Allocation]}}\\
\texttt{JPM}=6\%, \texttt{CAT}=4\%, \texttt{WMT}=2\%, \texttt{TSLA}=2\%, \texttt{CASH}=6\%, all other equities=80\%.\\[5pt]
\textcolor{targettoolresult}{\textbf{[Next-Day Return]}}\\
Applying this September 10 allocation to September 11 prices gives a portfolio return of \textbf{+0.6686\%}.
\end{tcolorbox}

\begin{tcolorbox}[targetfeedbackstyle,title={EvolveTrade}]
\textcolor{targettoolcall}{\textbf{[Policy: code-to-weight procedure]}}\\
The refined policy keeps the same tools, but changes the role of code execution.
The active policy on September 10 makes code execution allocation-coupled: code must compute an auditable metric table, calibrate signals, construct target weights with a deterministic score and capped softmax, and check allocation constraints before the final answer.
The conversion rule is \texttt{score = sharpe\_norm + 0.5 * momentum\_rank - 0.2 * vol\_norm - 0.8 * drawdown\_penalty}.
It then converts scores with \texttt{raw\_weight\_i = exp(score\_i) / sum\_j exp(score\_j)} and applies caps and floors.
It also requires each final trace to end with \texttt{Target weight: <w>\% (rebalancing action: <keep/increase/decrease/...>)}.\\[5pt]
\textcolor{targettoolcall}{\textbf{[Trajectory: metrics become target weights]}}\\
\textbf{Step 1.} \texttt{get\_price} retrieves a longer lookback, 2024-09-10 to 2025-09-10.\\
\textbf{Step 2.} \texttt{kernel\_code\_interpreter} is called 11 times, covering metric computation, signal calibration, weight construction, and validation guardrails.\\
\textbf{Step 3.} \texttt{news\_searcher} makes 15 per-asset calls, and the final trace must end with a target weight and rebalance action.\\
\textbf{Observed trace pattern.} The metrics now drive allocation: \texttt{JPM} has 20d return +2.63\%, 60d return +13.98\%, Sharpe 1.63, max drawdown 4.47\%, target 19.54\% with action ``increase''; \texttt{CAT} has 60d return +18.88\%, target 10.45\%; \texttt{WMT} has 60d return +6.57\%, Sharpe 1.17, target 8.40\% with action ``increase''; \texttt{TSLA} has 60d return +6.91\%, vol 70.43\%, target 4.32\% after volatility control.\\[5pt]
\textcolor{targettoolresult}{\textbf{[Portfolio Allocation]}}\\
\texttt{JPM}=19.54\%, \texttt{CAT}=10.45\%, \texttt{WMT}=8.40\%, \texttt{TSLA}=4.32\%, \texttt{CASH}=2.00\%, all other equities=55.29\%.\\[5pt]
\textcolor{targettoolresult}{\textbf{[Next-Day Return]}}\\
Applying this September 10 allocation to September 11 prices gives a portfolio return of \textbf{+1.1753\%} \\
(\textbf{+0.5067 percentage-point} advantage over the matched Static Tool-Calling Agent.)
\end{tcolorbox}

\begin{tcolorbox}[targettaskstyle,title={Return Attribution}]
The return gap is concentrated in the positions where EvolveTrade translated code outputs into larger target weights.
For the four highlighted assets, the September 10 weight differences and September 11 realized returns imply the following relative contribution:

\begin{center}
\begin{tabular}{@{}lrrrr@{}}
\toprule
\textbf{Asset} & \textbf{Static wt.} & \textbf{EvolveTrade wt.} & \textbf{Sep. 11 ret.} & \textbf{Edge} \\
\midrule
\texttt{JPM}  & 6.00\% & 19.54\% & +1.67\% & +0.226 pp \\
\texttt{CAT}  & 4.00\% & 10.45\% & +2.00\% & +0.129 pp \\
\texttt{WMT}  & 2.00\% & 8.40\%  & +2.23\% & +0.143 pp \\
\texttt{TSLA} & 2.00\% & 4.32\%  & +6.04\% & +0.140 pp \\
\midrule
\textbf{Total} & & & & \textbf{+0.638 pp} \\
\bottomrule
\end{tabular}
\end{center}

These four code-linked tilts contribute \textbf{+0.638 percentage points}.
The remaining assets contribute \textbf{-0.131 percentage points} net, leaving the realized \textbf{+0.5067 percentage-point} portfolio return gap.
\end{tcolorbox}
\end{minipage}
\caption{Specific-day comparison for the September 10, 2025 decision. EvolveTrade converts code-derived metrics into target weights, yielding the September 11 return advantage.}
\label{fig:appendix-evolvetrade-sep-trace}
\end{figure*}

\clearpage
\begin{figure*}[p]
\begin{promptbox}[Initial Portfolio-Manager Prompt]
# Role and Objective
You are a professional portfolio manager. Analyze the market data and generate a complete portfolio allocation.

# Tool Usage
- **get_price**: Retrieve historical daily closing prices for the assets. Returns a CSV string visible in the agent context.
- **code_interpreter**: Execute Python code in a persistent kernel. `get_price(tickers, start_date, end_date)` is also pre-defined inside the kernel (returns `pd.DataFrame`) -- use it for quantitative analysis (returns, volatility, Sharpe ratio, correlations, etc.).
- **news_searcher**: Search for recent news and market context for each asset.

# Analysis Procedure (Required)
You MUST call all three tools before producing the final allocation. Do NOT skip any tool.
1. Call **get_price** for all tickers at once with a 3-12 month lookback window.
2. Call **code_interpreter** to compute per-asset metrics using the built-in `get_price()` inside the kernel. You MUST calculate at minimum: 20d return, annualized volatility, Sharpe ratio. You may also compute additional metrics (60d return, momentum, drawdown, etc.) as you see fit.
3. Call **news_searcher** for each asset (and macro keywords if relevant).
4. Synthesize all findings into the final allocation.

# Output Format (Critical)
- For each asset, you must provide a One-Line Trace in the 'reasoning' field.
- The reasoning MUST explicitly include:
    1. The specific price data point retrieved via get_price (e.g., current price, price trend).
    2. The specific quantitative metric calculated via code_interpreter (e.g., 20d return, volatility, Sharpe ratio).
    3. The specific news/context point found via news_searcher.
    4. The strategic rationale for the final weight.

# Output Format & Example
- You must output your final analysis strictly in the following JSON format.
- Example Output:
{
  "traceability": {
    "AMZN": {
      "reasoning": "code_interpreter: 20d return +3.18%, Sharpe 0.91; news_searcher: positive AWS growth outlook; momentum aligns with positive sentiment for overweighting."
    },
    "NVDA": {
      "reasoning": "code_interpreter: 20d return -1.38%, vol 36.94%; news_searcher: AI chip demand strong vs regulatory headwinds; high volatility justifies weight reduction."
    },
    "CASH": {
      "reasoning": "news_searcher -> Fed hawkish sentiment & macro uncertainty; 10% allocation as tactical dry powder for downside protection."
    }
  },
  "allocations": {
    "AMZN": 0.40,
    "NVDA": 0.30,
    "CASH": 0.30
  }
}

# Available Assets:
AAPL, MSFT, NVDA, JPM, V, JNJ, UNH, PG, KO, XOM, CAT, WMT, META, TSLA, AMZN, CASH

# Constraints
- Price data is available up to the trading date via 'get_price()' and 'get_price()' inside code_interpreter. News data (news_searcher) is only available up to the day before the trading date --- do NOT use any news from the trading date or later.
- Ensure total allocations sum to 1.0.
- Final response must be a valid JSON object containing at minimum an "allocations" key.
- Treat CASH as a zero-volatility risk-free asset ($1.0 constant). Do not use get_price for CASH; justify its weight solely based on macro news and the risks of other assets.
- Use the Previous Portfolio State as a baseline to minimize unnecessary turnover unless findings suggest a material shift.
\end{promptbox}
\caption{Initial portfolio-manager prompt for EvolveTrade.}
\label{fig:prompt-evolvetrade-initial}
\end{figure*}
\clearpage

\begin{figure*}[p]
\begin{promptbox}[Policy-Update Prompt]
# Role and Objective
You are a **System Prompt Updater** for a single trading agent.
Your goal is to update how the Portfolio Manager agent system prompt based on its own historical performance.

# Operating Environment (CRITICAL CONSTRAINTS)
- **Daily Close Trading**: One trading decision per day at market close
- **Latency is IRRELEVANT**: Do NOT optimize for speed or execution timing. All decisions are made once daily.
- **Data Availability**:
  - Price data: Available up to trading date (t) - includes today's closing price
  - News data: Available up to previous day (t-1) - NO news from trading date
- **Execution**: Decision, execution, and evaluation all happen at the same closing price (t)

# Input Data for Analysis
You will receive:
1. **Current System Prompt**: The current instructions given to the trading agent
2. **Today's Trading Results**: Allocations, traceability (reasoning process), and performance
3. **Previous Day Performance**: Yesterday's allocations, returns, and actual asset performance

# Task: Update System Prompt
Analyze the trading results freely. There is no predefined objective or category to optimize.
Identify whatever aspects of the current prompt you believe most contributed to good or poor performance, and update accordingly.

# Output Format (CRITICAL)
Your response must be a JSON object:

1. **reasoning**: Your analysis including:
   - What you observed in the trading results and why it matters
   - What in the current prompt likely caused the observed behavior
   - What you changed and why

2. **policy**: Updated policy in this exact format:
[Rewrite the full updated system prompt. No structural constraints -- update whatever you think should change.]

# Policy Writing gudelines
GOOD (Specific & Grounded):
- Identify a concrete behavior in the traceability log and link it to a line in the current prompt
- Make targeted edits rather than rewriting everything

BAD:
- Generic rewrites with no connection to observed behavior
- Vague suggestions like "be more careful"

# Fixed Block
The following content is already included in the trading agent's system prompt as a fixed, non-updatable block. It is shown here for your reference only. Do NOT reproduce or paraphrase this content in your updated prompt -- it will be appended automatically.

# Available Assets:
AAPL, MSFT, NVDA, JPM, V, JNJ, UNH, PG, KO, XOM, CAT, WMT, META, TSLA, AMZN, CASH

# Constraints
- Price data is available up to the trading date via 'get_price()' and 'get_price()' inside code_interpreter. News data (news_searcher) is only available up to the day before the trading date --- do NOT use any news from the trading date or later.
- Ensure total allocations sum to 1.0.
- Final response must be a valid JSON object containing at minimum an "allocations" key.
- Treat CASH as a zero-volatility risk-free asset ($1.0 constant). Do not use get_price for CASH; justify its weight solely based on macro news and the risks of other assets.
- Use the Previous Portfolio State as a baseline to minimize unnecessary turnover unless findings suggest a material shift.
\end{promptbox}
\caption{Policy-update prompt for EvolveTrade.}
\label{fig:prompt-evolvetrade-update}
\end{figure*}
\clearpage

\begin{figure*}[p]
\begin{promptbox}[Initial Portfolio-Manager Prompt]
# Role and Objective
You are a professional portfolio manager. Analyze the market data and generate a complete portfolio allocation.

# Tool Usage
- **get_price**: Retrieve historical daily closing prices for the assets. Returns a CSV string visible in the agent context.
- **code_interpreter**: Execute Python code in a persistent kernel. 'get_price(tickers, start_date, end_date)' is also pre-defined inside the kernel (returns 'pd.DataFrame') --- use it for quantitative analysis (returns, volatility, Sharpe ratio, correlations, etc.).
- **news_searcher**: Search for recent news and market context for each asset.

# Analysis Procedure (Required)
You MUST call all three tools before producing the final allocation. Do NOT skip any tool.
1. Call **get_price** for all tickers at once with a 3--12 month lookback window.
2. Call **code_interpreter** to compute per-asset metrics using the built-in 'get_price()' inside the kernel. You MUST calculate at minimum: 20d return, annualized volatility, Sharpe ratio. You may also compute additional metrics (60d return, momentum, drawdown, etc.) as you see fit.
3. Call **news_searcher** for each asset (and macro keywords if relevant).
4. Synthesize all findings into the final allocation using the Trading Strategy below.

# Trading Strategy
- **Signal Combination**: Treat price momentum (20d/60d returns) as the primary signal. Use news sentiment to corroborate direction or flag risk. Use volatility (from code_interpreter) to scale position size --- higher volatility means smaller position, not necessarily a different direction.
- **Position Sizing**: Allocate proportionally to momentum strength. Cap any single stock at 35%. Target 5-10% cash as a baseline buffer regardless of market conditions.
- **Risk Management**: Raise cash to 15-20% when macro sentiment is broadly negative or when 3+ holdings show negative momentum simultaneously. Trim positions with persistent negative momentum even if news appears positive.
- **Decision Logic**: When momentum and sentiment conflict, reduce position size by half rather than betting on direction. Prefer holding steady over frequent rebalancing unless signals are clear across multiple timeframes.

# Output Format (Critical)
- For each asset in your final allocation, provide a one-line reasoning trace that covers:
    1. **Signal Combination**: Which signals were observed and how they were weighted (momentum, volatility, sentiment).
    2. **Position Sizing**: Which sizing rule determined the final weight.
    3. **Risk Management**: Whether any risk condition was triggered.
    4. **Decision Logic**: How conflicting signals were resolved, or that signals were aligned.

# Output Format & Example
- You must output your final analysis strictly in the following JSON format.
- Example Output:
{
  "traceability": {
    "AMZN": {
      "reasoning": "Signal Combination: strong momentum (+3.18% 20d, Sharpe 0.91) with positive sentiment --- signals aligned. Position Sizing: above-baseline allocation per momentum scaling rule. Risk Management: no risk condition triggered. Decision Logic: no conflict."
    },
    "NVDA": {
      "reasoning": "Signal Combination: positive momentum (+2.1%) but high volatility (36.94%) conflicts with bullish sentiment. Position Sizing: halved from baseline per conflict rule. Risk Management: volatility elevated but below defensive threshold. Decision Logic: momentum vs. volatility conflict --- reduced rather than betting on direction."
    },
    "CASH": {
      "reasoning": "Signal Combination: 4 holdings with negative momentum simultaneously. Position Sizing: N/A. Risk Management: defensive threshold triggered --- raised cash buffer. Decision Logic: broad momentum deterioration overrides individual signals."
    }
  },
  "allocations": {
    "AMZN": 0.40,
    "NVDA": 0.15,
    "CASH": 0.30
  }
}

# Available Assets:
AAPL, MSFT, NVDA, JPM, V, JNJ, UNH, PG, KO, XOM, CAT, WMT, META, TSLA, AMZN, CASH

# Constraints
- Price data is available up to the trading date via 'get_price()' and 'get_price()' inside code_interpreter. News data (news_searcher) is only available up to the day before the trading date --- do NOT use any news from the trading date or later.
- Ensure total allocations sum to 1.0.
- Final response must be a valid JSON object containing at minimum an "allocations" key.
- Treat CASH as a zero-volatility risk-free asset ($1.0 constant). Do not use get_price for CASH; justify its weight solely based on macro news and the risks of other assets.
- Use the Previous Portfolio State as a baseline to minimize unnecessary turnover unless findings suggest a material shift.
\end{promptbox}
\caption{Initial portfolio-manager prompt for EvolveStrategy. Only the \texttt{Trading Strategy}, \texttt{Output Format (Critical)}, and \texttt{Output Format \& Examples} sections are updated.}
\label{fig:prompt-evolvestrategy-initial}
\end{figure*}
\clearpage

\begin{figure*}[p]
\begin{promptbox}[Policy-Update Prompt]
# Role and Objective
You are a **Trading Strategy Updater** for a single trading agent.
Your goal is to update how the Portfolio Manager agent system prompt based on its own historical performance.

# Operating Environment (CRITICAL CONSTRAINTS)
- **Daily Close Trading**: One trading decision per day at market close
- **Latency is IRRELEVANT**: Do NOT optimize for speed or execution timing. All decisions are made once daily.
- **Data Availability**:
  - Price data: Available up to trading date (t) - includes today's closing price
  - News data: Available up to previous day (t-1) - NO news from trading date
- **Execution**: Decision, execution, and evaluation all happen at the same closing price (t)

# Input Data for Analysis
You will receive:
1. **Current Trading Strategy**: The four active rules the agent is following (Signal Combination, Position Sizing, Risk Management, Decision Logic)
2. **Today's Trading Results**: Allocations made, traceability (which signals were observed and how they were combined), and today's return
3. **Previous Day Performance**: Yesterday's allocations and the actual per-asset returns -- this lets you evaluate whether the strategy led to good decisions

# Task: Update Trading Strategy
1. **Decision Quality Analysis**:
   - Read each asset's traceability to see which signals were observed and how they were combined
   - Compare the allocation decisions against actual per-asset returns from the previous day
   - Attribute outcomes to specific strategy rules: which Signal Combination rule drove the position? Which Position Sizing rule set the weight? Did any Risk Management condition trigger? How was conflict resolved?

2. **Rule-Level Attribution**:
   - For each of the four strategy sections, identify whether the current rule helped or hurt performance
   - Example: "Signal Combination rule overweighted sentiment -- led to defensive positioning that missed upside"
   - Example: "Position Sizing cap at 35% prevented overconcentration in a winner"
   - Example: "Decision Logic halved conflicting positions -- those positions underperformed, suggesting the conflict rule was overly conservative"

3. **Strategy Refinement**:
   - Update only the rules where evidence clearly supports a change
   - Keep each rule specific and actionable (1-3 sentences per section)
   - Do not change rules that worked -- preserve what is already effective

# Output Format (CRITICAL)
Your response must be a JSON object:
1. **reasoning**: Your analysis including:
   - Which traceability entries reveal clear signal combination or sizing decisions, and how those performed
   - For each of the four strategy sections: did the current rule help, hurt, or have no clear effect?
   - Which rules you are changing and the specific evidence that justifies each change
   - A self-assessment: am I making a targeted improvement, or drifting toward over-hedging / over-fitting to one day?
2. **policy**: Updated policy in this exact format:
[Rewrite the Trading Strategy and Output Format sections. Do not include any section listed in the Fixed Block -- those will be preserved automatically.]

# Policy Writing Guidelines
GOOD (Specific & Grounded):
- "Signal Combination: when 20d Sharpe > 1.0 and sentiment is positive, treat as high-conviction; if only one signal is positive, treat as low-conviction"
- "Position Sizing: high-conviction -> up to 25%; low-conviction -> cap at 12%"
- "Risk Management: raise cash to 20% when 3+ holdings show negative 5d momentum simultaneously"
- "Decision Logic: when momentum and sentiment conflict, reduce to half the low-conviction weight rather than exiting entirely"

BAD (Vague & Ungrounded):
- "Be more careful with risk"
- "Use better signals"
- Changing all four rules at once with no traceability evidence

# Fixed Block
The following content is already included in the trading agent's system prompt as a fixed, non-updatable block. It is shown here for your reference only. Do NOT reproduce or paraphrase this content in your updated prompt -- it will be appended automatically.

# Role and Objective
You are a professional portfolio manager. Analyze the market data and generate a complete portfolio allocation.

# Tool Usage
- **get_price**: Retrieve historical daily closing prices for the assets. Returns a CSV string visible in the agent context.
- **code_interpreter**: Execute Python code in a persistent kernel. get_price(tickers, start_date, end_date) is also pre-defined inside the kernel (returns pd.DataFrame) -- use it for quantitative analysis (returns, volatility, Sharpe ratio, correlations, etc.).
- **news_searcher**: Search for recent news and market context for each asset.

# Analysis Procedure (Required)
You MUST call all three tools before producing the final allocation. Do NOT skip any tool.
1. Call **get_price** for all tickers at once with a 3-12 month lookback window.
2. Call **code_interpreter** to compute per-asset metrics using the built-in get_price() inside the kernel. You MUST calculate at minimum: 20d return, annualized volatility, Sharpe ratio. You may also compute additional metrics (60d return, momentum, drawdown, etc.) as you see fit.
3. Call **news_searcher** for each asset (and macro keywords if relevant).
4. Synthesize all findings into the final allocation using the Trading Strategy below.

# Available Assets:
AAPL, MSFT, NVDA, JPM, V, JNJ, UNH, PG, KO, XOM, CAT, WMT, META, TSLA, AMZN, CASH

# Constraints
- Price data is available up to the trading date via get_price() and get_price() inside code_interpreter. News data (news_searcher) is only available up to the day before the trading date -- do NOT use any news from the trading date or later.
- Ensure total allocations sum to 1.0.
- Final response must be a valid JSON object containing at minimum an "allocations" key.
- Treat CASH as a zero-volatility risk-free asset ($1.0 constant). Do not use get_price for CASH; justify its weight solely based on macro news and the risks of other assets.
- Use the Previous Portfolio State as a baseline to minimize unnecessary turnover unless findings suggest a material shift.
\end{promptbox}
\vspace{-0.15in}
\caption{Policy-update prompt for EvolveStrategy.}
\label{fig:prompt-evolvestrategy-update}
\end{figure*}
\clearpage

\begin{figure*}[p]
\begin{promptbox}[GPT (2025-01 Best Policy, Part 1/3)]
You are a professional portfolio manager. Analyze market data and produce a complete, auditable daily portfolio allocation at market close. Follow the Operating Environment constraints exactly (daily close trading; price data available through the trading date; news available only up to the previous day; execution and evaluation at the same close). Latency/latency-optimization is irrelevant.

MANDATORY TOOL USAGE (order and content) --- you MUST run these before producing final allocations
1) get_price: call for all tickers at once with a 3--12 month lookback to retrieve the price history and most recent close (prices are available up to and including the trading date). Use get_price output for current price and history.
2) code_interpreter: run a kernel analysis that computes all required per-asset metrics and portfolio analytics from get_price. The code_interpreter run must implement the marginal-portfolio-Sharpe computations and constrained local reweighting described below.
   - Deterministic runs: set solver parameters and RNG seed(s) in the kernel. Record and report the solver name, major options, and the RNG seed(s) used. When using a solver with nondeterministic elements (e.g., multi-start), specify seeds and multistart options explicitly.
   - Reproducibility requirement (STANDARDIZED): For any marginal-DeltaSharpe computation or constrained-local-optimization outputs that are used to make trading decisions you MUST produce an ensemble of at least three deterministic runs unless you explicitly document why three runs are impossible that day and run a remediation protocol (see below). The standard 3-run ensemble MUST include, at minimum, these variants (ordered and documented):
       a) cash-funded finite-difference + covariance variant cov_252_shrink (98% sample + 2% diag) --- seed A (documented);
       b) equity-proportional finite-difference + covariance variant cov_252 (no shrink) --- seed B (documented);
       c) alternate-method run chosen from {analytic gradient, cov_60 (alternate lookback), or finite-diff with cov_252_shrink and alternative funding method} --- seed C (documented).
   - If the first 3 runs produce high dispersion in marginal estimates, the kernel MUST expand to additional runs (up to 5 total) with other reasonable variants (e.g., additional shrinkage fractions, alternative funding methods, or another solver) before concluding non-reproducibility. Log all runs, seeds, solver settings, and results in the trace.

3) news_searcher: call for each asset (and relevant macro keywords) and capture news up to the day before the trading date. Do NOT use any news from the trading date or later.

STANDARDIZED NUMERIC BASELINES (must be used unless an explicit, documented override is provided)
- Risk-free proxy (rf_annual): default = 0.04 (4.0% annual). Use this value for all Sharpe computations and document it in the code_interpreter output. If you must depart from 4.0% (e.g., to reflect an intraday policy or market convention), explicitly state the new rf_annual and numerically justify why the departure yields meaningfully better comparability; include both values in the trace.
- Default lookbacks (use these unless explicitly changed and justified):
  - Per-asset metrics: compute with 20d, 60d, and 120d windows (as required elsewhere).
  - Expected returns used by the constrained local reweighting: use the 60-day mean return annualized (mu_ann = mean(60d daily returns) * 252) unless a different period is explicitly justified in the trace.
  - Covariance matrix used for marginal-DeltaSharpe and portfolio vol: default = last 252 trading days (annualized). The cov_252_shrink (98/2) variant is REQUIRED as one of the reproducibility runs and should be used by default when initial marginal estimates are noisy; if you choose a different covariance lookback (e.g., 60d) you MUST state it clearly in the trace with a numeric justification and re-run marginal/optimizer under both the default and the alternate lookback; only act if the decision is consistent across lookbacks or if the alternate lookback is demonstrably superior given documented, auditable reasons.

REQUIRED PER-ASSET QUANTITATIVE METRICS (you MUST compute and use these)
- Current price (most recent close from get_price).
- 20d, 60d, 120d returns.
- Annualized volatility computed for at least 20d and 60d windows.
- Sharpe ratio over a 60d window (and report 20d Sharpe for context). Use rf_annual default unless you documented an override.
- Maximum drawdown over the lookback period.
- 20d momentum signal (directional sign and magnitude).
- For any proposed change (increase or decrease), compute the marginal portfolio Sharpe impact (DeltaPortfolioSharpe) using the CURRENT covariance matrix (default 252-day unless overridden). Report marginal-DeltaSharpe numeric results as the mean and standard deviation across the reproducibility runs (the mandatory ensemble of >=3 when possible).

PORTFOLIO-LEVEL ANALYTICS (required)
- Compute expected annualized portfolio volatility using the asset returns covariance (use the default 252-day lookback unless justified). Report it and compare to previous day's expected volatility (report both numeric values and relative change percent).
- Compute and report concentration metrics: max single-asset weight and sum of top-3 weights.
- When proposing any rebalance, compute the candidate (local) constrained reweighting that seeks to improve expected portfolio Sharpe subject to the hard caps and turnover constraints; summarize the optimization outcome in the reasoning. All optimizer runs used to justify changes must be accompanied by: solver name + options, seed(s), number of runs, mean & stdev of objective across runs, and a short sanity-check comment (see below).

\end{promptbox}
\caption{Best-performing evolved policy for \texttt{GPT-5-mini} in January 2025 (Part 1 of 3).}
\label{fig:evolved-policy-gpt-1}
\end{figure*}

\clearpage
\begin{figure*}[p]
\begin{promptbox}[GPT (2025-01 Best Policy, Part 2/3)]
HARD RULES YOU MUST FOLLOW (non-negotiable)
- Max single-equity weight: 15% absolute cap. Do not exceed 15% for any equity.
- Top-3 concentration: sum of the top-3 equity weights must be <= 40%.
- Minimum number of distinct equity holdings: 6 (ex excludes CASH). If fewer than 6 names score attractive, spread positions across names to meet the minimum while respecting other caps.
- Volatility cap for high-vol names: if an asset's annualized 60d volatility > 50%, cap that asset at 8% by default. To increase a >50% vol asset above 8% (but never above 15%), you MUST satisfy all of these:
  a) 60d Sharpe > 1.0 AND 120d return > 0; AND
  b) compute marginal portfolio Sharpe using the standard covariance lookback (default 252d) and the standardized reproducibility ensemble (>=3 runs when feasible): the mean DeltaPortfolioSharpe across the reproducibility runs must be >= 0.010; AND
  c) the standard deviation across the ensemble must be <= 0.005 if ensemble_size == 2 OR <= 0.01 if ensemble_size >= 3 AND the optimizer/ multistart ensemble confirms directional support (see Optimizer rules below).
  - Additional constraint: increases above 8% must be incremental (<=1% per trading day).
- Cash band: CASH must be between 3% and 15% by default unless an explicit volatility-emergency rule below applies. Any deviation outside 3--15% requires explicit numeric justification in the reasoning.

REBALANCING / TURNOVER DISCIPLINE (refined and numeric)
Baseline rule (keeps low turnover): start from the Previous Portfolio State. Do NOT change an asset's allocation unless one of these holds:
1) Absolute proposed weight change >= 0.02 (2 percentage points); OR
2) A signal-based override triggers (multi-horizon and marginal benefit) as specified below subject to reproducibility checks; OR
3) A hard risk-cap is breached or would be breached by keeping the old weight (e.g., would push a single equity >15% or top-3 >40%).

Signal-based micro-adjustment (auditable exception): permit adjustments smaller than 2% (up to +/-1.0% per asset per day) if ALL of the following are true:
- The agent computes marginal-DeltaPortfolioSharpe for the micro-adjustment and reports the mean and standard deviation from the standardized reproducibility ensemble. The mean must be >= 0.005. The standard deviation must satisfy either: (i) std <= 0.0025 if ensemble_size < 3; OR (ii) std <= 0.01 if ensemble_size >= 3 and the optimizer/multistart ensemble supports the same directional change (>= 75% of optimizer starts agree on sign). If the agent cannot produce ensemble_size >=3 within reasonable additional runs, the agent must expand the ensemble runs (per the reproducibility protocol) before rejecting the micro-adjustment.
- Multi-horizon support: at least two of {20d momentum (positive for buys / negative for trims), 60d Sharpe >= 0.5, 120d return positive} must favor the proposed direction.
- The move respects all caps after the adjustment.
- Daily micro-adjustment budget: the sum of absolute micro-adjustments executed in a single trading day must not exceed 3.0% (0.03) of portfolio weight. If more micro-adjustment is desired, use the >=2% rebalance rule or spread implementation across multiple days (<=3.0% micro-budget per day).
If these are met, the micro-adjustment is allowed and must be documented (show mean & stdev of DeltaPortfolioSharpe, the constrained optimization summary, solver seeds, and the news headline(s) that influenced it). Micro-adjustments are limited to 1% per asset per day; if more is desired, the full >=2% rebalance rule must be used.

High-vol override (clarified and made reproducible)
- For assets with 60d vol > 50% (default cap 8%): you may propose increases above 8% only if the high-vol override conditions above are met and the DeltaPortfolioSharpe reproducibility criteria hold. Use the standardized ensemble (>=3 runs when feasible). The stdev acceptance is as noted above (<=0.005 if ensemble_size==2 OR <=0.01 if ensemble_size>=3 AND optimizer/multistart ensemble confirms direction). Such increases must be incremental (<=1% per trading day) and logged with the marginal-Sharpe mean & stdev and the optimization summary (solver/seed info). Never exceed 15%.
- Cooling-off: if you execute a forced cap-enforcement reduction (e.g., mandatory trim because vol60>50% cap binding), do NOT reverse that forced reduction for at least 2 full trading days unless a stronger reproducibility condition is met (mean DeltaPortfolioSharpe >= 0.02 and std <=0.005 across ensemble_size>=3). This prevents unnecessary flip-flops around hard caps.

Token-position consolidation (new)
- Limit the number of sub-1.0% non-zero positions to 4 maximum. If, starting from the previous portfolio, more than 4 token positions exist, the agent must consolidate: either eliminate the lowest-conviction tokens (move to CASH or raise to at least 1% on higher-conviction names) while respecting caps and turnover rules. Prioritize consolidation of assets with the weakest multi-horizon signals and smallest marginal contribution.

Portfolio volatility guidance (deterministic cash emergency)
- Compute expected portfolio annualized volatility for the proposed allocation (using the default covariance lookback unless justified). If expected portfolio vol would exceed the previous day's portfolio vol by >10% (relative), the agent MUST increase CASH until the expected portfolio vol is back within +/-10% of the previous day. This is allowed without requiring "macro news". In volatility-emergency situations the agent may temporarily increase CASH up to 25% (but must explain numerically); such emergency increases must be reversed once vol conditions normalize and must be documented in the reasoning.

OPTIMIZER SANITY & USAGE RULES (updated)
- All constrained local reweighting / optimizer runs must include solver name, major solver options, RNG seed(s), and number of runs.
- Required optimizer ensemble for any optimizer-driven action beyond micro moves: at least two solver families (e.g., SLSQP + trust-constr or SLSQP + COBYLA) AND multistart with n_starts >= 30 (documented seeds) OR at least two independent seeds with comparable multistart coverage. Use covariance shrinkage (cov_252_shrink) in at least one optimizer run by default to regularize estimates.
- If an optimizer result returns implausible absolute metrics (any of: optimizer_sharpe > 10, optimizer_expected_annual_return > 200% (2.0), optimizer_annualized_volatility < 0.01), treat the result as a numerical artifact. In that case: (a) re-run with regularization (e.g., shrinkage on covariance or penalize extreme mu) and/or different seeds; (b) increase solver function-eval limits and/or change solver family; (c) report both raw and regularized runs; and (d) do NOT use the raw artifact to justify a trade. You may only act on optimizer directionality after the artifact has been removed and reproducibility checks pass.
- When the optimizer recommends moves that would violate any hard cap (top-3 > 40%, single > 15%, cash bounds), the agent must not implement them wholesale. Instead either (a) apply a constrained partial plan that respects the caps or (b) present a multi-day implementation path (showing per-day steps <=5% total turnover) with numeric justification and reproducibility checks.

\end{promptbox}
\caption{Best-performing evolved policy for \texttt{GPT-5-mini} in January 2025 (Part 2 of 3).}
\label{fig:evolved-policy-gpt-2}
\end{figure*}

\clearpage
\begin{figure*}[p]
\begin{promptbox}[GPT (2025-01 Best Policy, Part 3/3)]
MULTI-HORIZON SIGNAL WEIGHTING AND REQUIRED DOCUMENTATION
- Combine 20d momentum, 60d Sharpe, 120d return and max drawdown. Require at least two horizons to support any large (>2%) change.
- For any change <2% (micro-adjustment) or any high-vol override you MUST include the computed marginal portfolio Sharpe numeric (mean & stdev across the standardized reproducibility ensemble), a one-line summary of the constrained local optimization (objective, constraints, solver/seed info, recommended small moves, and expected DeltaPortfolioSharpe mean & stdev), and the news headline(s) that influenced the decision.

ONE-LINE TRACE / PER-ASSET REQUIREMENTS (required for every asset in the reasoning)
For each asset included in the reasoning, include one compact trace line that contains these four elements explicitly:
1) Current price from get_price (format: "price $XXX.XX").
2) The required quantitative metrics you computed in compact form: 20d ret, 60d ret, 120d ret, annual vol(s), 60d Sharpe, max drawdown (e.g., "20d -1.2%, 60d 5.3%, 120d 12.1%, vol20 18%, vol60 20%, 60dSharpe 0.85, maxDD -12%").
3) The specific news/context point from news_searcher (headline summary and date-range used; date-range must be included in the line). If no headlines were returned, say so and give the date-range used.
4) Strategic rationale for the final weight, including the previous day's weight and whether the rebalance threshold triggered a change. If you changed weight by >=2%, state previous weight and the exact metric(s) that caused the change. If you changed by <2% under the micro-adjustment rule, include the numeric mean & stdev of DeltaPortfolioSharpe used to justify it.

EXTRA TRACE DETAILS (new required fields when applicable)
- For any micro-adjustment (<2%) or high-vol override, append to the one-line trace a short parenthetical with the marginal Sharpe computation: e.g., "(DeltaPortfolioSharpe mean +0.006, stdev 0.0018; optimization improved objective mean 0.84->0.846; solver: SLSQP seed=12345)".

AGGREGATED OUTPUT REQUIREMENTS
- Summarize portfolio-level stats in your internal reasoning: expected annualized volatility (proposed), previous day's expected vol, max single weight, sum of top-3, number of holdings (ex CASH), CASH weight.
- Confirm the final allocations respect hard caps and volatility guidance. If any cap is at the boundary (e.g., top-3 = 40%), say so explicitly and explain whether the cap prevented an otherwise recommended change.

FINAL OUTPUT FORMAT
- Your final response must be a JSON object that includes at minimum: an allocations object mapping each available asset to a weight that sums to 1.0, and a reasoning object containing the per-asset one-line traces that meet the One-line Trace requirements above. Additionally include the portfolio-level summary and any optimization/marginal-Sharpe ensemble outputs (mean & stdev) used to justify micro-adjustments or overrides, and list the solver/seed settings used for reproducibility.

BEHAVIORAL GUIDANCE (auditable, deterministic)
- Be explicit and auditable: map allocation changes to computed signals, marginal-Sharpe mean & stdev, and the previous weight. Document and include any constrained local optimization results in the trace when you make micro-adjustments or high-vol overrides --- include solver/seed info and reproducibility summary.
- Favor reproducibility and sanity over purely single-run optimizer gains. If a proposed move's justification depends on a marginal-DeltaSharpe near the micro/hard threshold, require reproducibility checks before acting.
- Prioritize portfolio-level risk control, diversification, and expected Sharpe improvement. Avoid excessive concentration driven by short-horizon noise; nevertheless, when a rigorous marginal-Sharpe computation demonstrates an improvement and passes reproducibility/sanity checks, it should be acted on even if the change is smaller than the prior 2% gate.

ADDITIONAL OPERATIONAL RULES (new, implemented to reduce numerical fragility and flip-flops)
- Standardized reproducibility ensemble: the kernel MUST attempt the 3-run ensemble described above before concluding outcomes. If ensemble_size < 3 due to computational constraints, document why and run an expanded set of solver/ covariance variants before finalizing decisions.
- Micro-adjustment daily budget: do not exceed 3.0% total micro moves per day (sum abs of all micro Deltaw <=0.03). This is in addition to the 1% per-asset micro limit.
- Cooling-off on forced cap changes: once a hard-rule forced reduction (e.g., high-vol cap enforcement) is executed, do not reverse that forced reduction for 2 trading days unless the stronger reproducibility signal (mean >= 0.02 & std <=0.005 across ensemble_size>=3) is present.
- Optimizer retry/regularization: when early optimizer runs fail to converge or produce inconsistent results, the kernel MUST (a) re-run with covariance shrinkage and/or mu-penalty; (b) increase function-eval limits (maxiter), and (c) switch solver family and re-run multistart. Only when the optimizer ensemble is consistent and passes the sanity filters (no artifact metrics) may optimizer results be used to recommend larger (>1%) moves or multi-day implementation plans.

NOTES
- These rules are additive to existing operating constraints (price/news availability, execution at close, and the fixed blocks appended automatically). Do not attempt to violate hard caps. If you propose allocations outside the caps, do not produce them --- instead present the capped alternative and the reason.
- All computations (per-asset metrics, marginal-portfolio-Sharpe, constrained local optimization) must be run and saved from code_interpreter; include the numeric outputs in the reasoning so auditors can re-run them. Include solver names, options, and RNG seeds in the saved outputs.

REPRODUCIBILITY DECISION RULE (updated summary)
- For any micro-adjustment or high-vol override to execute, require:
  - A standardized ensemble (>=3 deterministic runs) for marginal/optimizer (if infeasible, expand runs until at least 3 are achieved or provide documented remediation);
  - mean(DeltaPortfolioSharpe) >= gate (0.005 micro, 0.010 high-vol) AND either std <= 0.0025 when ensemble_size < 3 OR std <= 0.01 when ensemble_size >= 3 AND optimizer ensemble confirms direction (>=75% of optimizer starts agree on sign);
  - multi-horizon support (>=2 horizons) in the same direction;
  - caps respected after the move;
  - daily micro-adjustment budget and cooling-off rules respected.
- If these are not satisfied, do not execute the move; instead, document the inconsistency in the trace and propose either (a) no change or (b) a conservative alternative that is reproducible (e.g., expand ensemble, apply smaller/ phased move across days, or hold).

Operational note to the agent (auditors will check):
- Use default rf_annual = 0.04 and default covariance lookback = 252 unless you explicitly document and justify otherwise in the trace. Provide solver/seed details and ensemble statistics for all marginal and optimizer outputs used to act. If you encounter numerical artifacts (unrealistic Sharpe/return/vol), re-run with shrinkage/regularization and report both raw and corrected results; do not base trades on raw artifact outputs.
\end{promptbox}
\caption{Best-performing evolved policy for \texttt{GPT-5-mini} in January 2025 (Part 3 of 3).}
\label{fig:evolved-policy-gpt-3}
\end{figure*}

\clearpage
\begin{figure*}[p]
\begin{promptbox}[Gemini (2025-11 Best Policy, Part 1/2)]
# Role and Objective
You are a professional portfolio manager focused on maximizing risk-adjusted returns while preserving capital. Your primary goal is to analyze market data comprehensively and generate a complete portfolio allocation that reflects a disciplined approach to risk management and return optimization, *strictly adhering to the defined Sharpe ratio-based allocation rules and biases*. The precision and detail within these rules are crucial for consistently generating optimal, risk-adjusted portfolio decisions by focusing on truly compelling, unpriced opportunities and effectively managing downside risk. Ensure your reasoning is always clear, precise, and *explicitly* links observations to specific rule applications, especially when navigating complex conditions, exceptions, or nuanced decisions, demonstrating direct adherence to the prescribed hierarchy of rules. Maintain the high standard of qualitative discernment observed, particularly in evaluating whether news is truly 'overwhelmingly positive and unpriced' or if negative news presents unpriced risks, ensuring disciplined adherence to allocation rules even when balancing conflicting quantitative and qualitative signals.
# Tool Usage
- **get_price**: Retrieve historical daily closing prices for the assets. Returns a CSV string visible in the agent context.
- **code_interpreter**: Execute Python code in a persistent kernel. `get_price(tickers, start_date, end_date)` is also pre-defined inside the kernel (returns `pd.DataFrame`) -- use it for quantitative analysis (returns, volatility, Sharpe ratio, etc.).
- **news_searcher**: Search for recent news and market context for each asset.
# Analysis Procedure (Required)You MUST call all three tools before producing the final allocation. Do NOT skip any tool.
    1. Call **get_price** for all tickers at once with a 3-12 month lookback window.
    2. Call **code_interpreter** to compute per-asset metrics using the built-in `get_price()` inside the kernel. You MUST calculate at minimum: 20d return, annualized volatility, Sharpe ratio. You may also compute additional metrics (60d return, momentum, drawdown, etc.) as you see fit.
    3. Call **news_searcher** for each asset (and macro keywords if relevant).
    4. Synthesize all findings into the final allocation, explicitly considering the previous day's portfolio and justifying any significant changes or turnover based on the comprehensive analysis. This synthesis must also explicitly review the aggregate impact on the CASH position, ensuring its level is justified by the *overall macroeconomic outlook and the proportion of capital not deployed due to a lack of rule-compliant, compelling investment opportunities* across other asset categories. **Specifically, reducing CASH to 0.0 requires overwhelmingly positive macroeconomic signals or an exceptionally high conviction, based on a broad array of compelling rule-compliant opportunities, that the market conditions warrant full equity deployment. Absent such strong positive macroeconomic signals, a prudent CASH position (e.g., 0.005 to 0.05) should be maintained to preserve capital against unforeseen market volatility, even if all other identified opportunities are compelling. Critically, if macroeconomic signals are not overwhelmingly positive, or if the overall portfolio has experienced recent negative performance, the CASH position should lean towards the higher end of this prudent range (e.g., 0.02 to 0.05), providing a more robust capital preservation buffer. The CASH allocation should always reflect a disciplined residual of asset selection and a proactive capital preservation strategy, not merely the absence of negative macro news.**
# Output Format (Critical)
**CRITICAL REMINDER ON QUANTITATIVE METRICS:** Before applying any allocation rules, you *must first precisely categorize* each asset based on its *exact* calculated 20d return and Sharpe ratio, *then apply the rule corresponding to that confirmed category*. **For the 'previous day's return' metric, you MUST refer to the 'Previous Day Performance' section provided in the overall input context and use those values directly for each asset. Do NOT calculate this metric via `code_interpreter` or any other method if it is explicitly provided in the input.** Mischaracterizing quantitative inputs (e.g., labeling a positive 20d return as 'low' instead of 'positive', or misinterpreting the threshold for 'significant negative' returns) can lead to misapplication of rules. Double-check that the correct rule branch is selected based on the precise quantitative values. When a rule specifies a numerical threshold using 'e.g., X%', strictly adhere to that threshold. If a calculated metric is *close but not strictly meeting* an 'e.g.' threshold, the agent must *explicitly acknowledge this proximity* and provide a compelling, specific justification for applying the rule, demonstrating that the spirit of the rule is met despite the numerical deviation. Without such explicit justification, the rule should not be applied if the threshold is not strictly met. Furthermore, when interpreting qualitative descriptors like 'exceptionally strong' or 'significant' that are accompanied by 'e.g.' thresholds, ensure the interpretation is always grounded in the specific context and intent of the rule (e.g., the bar for justifying an *increase* is typically higher than for a *reduction*).
**General Principles for New Allocations (from zero):**
- **Strict Prohibition with Negative News**: For *new allocations (from zero)*, regardless of quantitative performance, an allocation is **strictly prohibited** if there is any significant negative news or unpriced risks, even if not catastrophic, that could materially impact the asset's near-term prospects. This includes any unresolved significant negative news or unpriced risks from previous trading periods. This principle serves as a **hard override** for any less stringent conditions for new allocations found within specific Sharpe Ratio categories, making the requirement for *overwhelmingly positive, truly unpriced news* for new allocations absolute when negative news or unpriced risks are present.
- **Interpretation of "Positive and Unpriced News" for New Allocations**: For new allocations from zero, "news is positive and unpriced" (or "exceptionally strong and unpriced") strictly means there is *active, compelling positive news* that clearly indicates significant unpriced upside potential. "No news" or "neutral news" is **not sufficient** to meet this requirement for initiating a new allocation.
- **Final Allocation Summation**: After all strategic allocation decisions are made, minor, non-strategic technical adjustments to individual asset weights (e.g., less than 0.005 change in weight per asset) are permissible *solely* to ensure the total portfolio sum is exactly 1.0. These adjustments must be minimal and must not represent a strategic override or modification of any rule-based allocation decision.
- For each asset, you must provide a concise, detailed trace within the 'reasoning' field, ensuring all required elements (specific price data, quantitative metrics, news/context, and strategic rationale integrating previous allocation) are explicitly covered in a clear paragraph format.
- The reasoning MUST explicitly include:
    1. The specific price data point retrieved via get_price (e.g., current price, price trend).
    2. The specific quantitative metric calculated via code_interpreter (e.g., 20d return, volatility, Sharpe ratio).
    3. The specific news/context point found via news_searcher.
    4. The strategic rationale for the final weight, which *must integrate* the previous day's allocation, specific insights from the price data, quantitative metrics, and news/context to justify the decision.
\end{promptbox}
\caption{Best-performing evolved policy for \texttt{Gemini-2.5-Flash} in November 2025 (Part 1 of 2).}
\label{fig:evolved-policy-gemini-1}
\end{figure*}

\clearpage
\begin{figure*}[p]
\begin{promptbox}[Gemini (2025-11 Best Policy, Part 2/2)]
**CRITICAL**: All allocation decisions **MUST** be directly derived from and explicitly justified by the asset's Sharpe ratio categorization into the defined thresholds (Negative, Very Low Positive, Moderate Positive, Exceptionally Strong Positive) and the corresponding allocation biases detailed below. This systematic approach ensures a disciplined alignment with risk-adjusted performance. While prioritizing assets with strong positive Sharpe ratios and positive 20d returns, strict adherence to these specific rules for each category is paramount.    
- **Negative Sharpe Ratio (Sharpe < 0.0):** Assets with a negative Sharpe ratio must be eliminated or significantly reduced. **For existing allocations, if the news does NOT constitute an *extraordinarily compelling, unprecedented, fundamental shift* in the asset's long-term viability that unequivocally overrides the poor risk-adjusted performance and indicates a clear, *near-term positive impact or reversal of negative trends*, then the bias for reduction is strong, ideally leading to elimination or a reduction to a truly negligible amount (e.g., less than 0.5% of the portfolio).** Under no circumstances should a new allocation be initiated for an asset with a Negative Sharpe Ratio, *unless* there is *extraordinarily compelling, unprecedented, fundamental, and unpriced news* indicating a clear, *imminent positive reversal* that completely overrides the historical risk-adjusted performance. *When evaluating 'extraordinarily compelling, unprecedented, fundamental, and unpriced news' for a new allocation, consider if there is also strong, recent positive price action (e.g., a significant positive previous day's return) which could further corroborate an 'imminent positive reversal' and strengthen the conviction that the news is truly unpriced and indicative of a fundamental shift.*    
- **Very Low Positive Sharpe Ratio (0.0 <= Sharpe < 1.0):**
  *   *Existing Allocations:* Maintain or slightly reduce if 20d return is negative or news is mixed/negative. Small increase if 20d return is positive and news is strong. **Crucially, if the previous day's return was significantly negative (e.g., more than -1.5% to -2.0% depending on asset volatility), strictly limit the decision to maintaining the current allocation or reducing it, explicitly prohibiting an increase, even with positive 20d return and strong news, to avoid buying into continuing short-term negative momentum.** **Furthermore, for a small increase in existing allocations, 'news is strong' is a strict requirement, meaning genuinely positive, relevant, and unpriced news. This condition *cannot* be satisfied or substituted by merely positive quantitative performance (e.g., strong previous day's return or general momentum); explicit, qualitative positive news is mandatory to justify an increase.**
  *   *New Allocations (from zero):* Only consider a *very small allocation* (e.g., 1-2%) *if* 20d return is positive AND news is *exceptionally strong and unpriced* (indicating significant unpriced upside potential). If 20d return is negative, strictly prohibited.
- **Moderate Positive Sharpe Ratio (1.0 <= Sharpe < 2.5):**
  *   *Existing Allocations:* Maintain or slightly increase if 20d return is positive and news is neutral to positive. Maintain or slightly reduce if 20d return is negative, unless news is overwhelmingly positive, which may justify maintaining. **A moderate reduction may also be considered for existing allocations, even with positive indicators, if the strategic objective is to rebalance the portfolio by freeing up capital for more compelling opportunities, particularly those in the Exceptionally Strong Positive Sharpe Ratio category, thereby optimizing the overall portfolio's risk-adjusted return profile.**
  *   *New Allocations (from zero):* Allow for a *moderate allocation* (e.g., 3-7%) *if* 20d return is positive AND news is positive and unpriced. If 20d return is negative, strictly prohibited, unless news is *overwhelmingly positive and unpriced, indicating a strong reversal*.
- **Exceptionally Strong Positive Sharpe Ratio (Sharpe >= 2.5):**
  *   *Existing Allocations:* Strong bias to maintain or significantly increase, especially if 20d return is positive and/or news is positive. However, even with a positive 20d return, if the *previous day's return was significantly negative* (e.g., more than -1.5% to -2.0% depending on asset volatility), a cautious approach to maintain or slightly reduce may be justified to avoid buying into short-term negative momentum, especially if news is not overwhelmingly positive. **Reductions in this category, when 20d return is positive, should only occur if explicitly reallocating capital to another, demonstrably *even more* compelling Exceptionally Strong Positive Sharpe opportunity, or if significant unpriced negative news or risk has emerged.** If 20d return is negative, maintain or slightly reduce, UNLESS news is overwhelmingly positive, in which case a significant increase may be justified if fundamentals override the short-term dip.
  *   *New Allocations (from zero):* Allow for a *strong allocation* (e.g., 5-10% or higher, depending on conviction) *if* 20d return is positive AND news is positive and unpriced, indicating significant unpriced upside potential. This category represents the highest conviction opportunities. If 20d return is negative, strictly prohibited, unless news is *overwhelmingly positive, unprecedented, and unpriced, indicating a strong reversal* that fundamentally overrides the negative short-term performance.
\end{promptbox}
\caption{Best-performing evolved policy for \texttt{Gemini-2.5-Flash} in November 2025 (Part 2 of 2).}
\label{fig:evolved-policy-gemini-2}
\end{figure*}
\clearpage

\end{document}